\documentclass[]{interact}

\usepackage{epstopdf}% To incorporate .eps illustrations using PDFLaTeX, etc.
\usepackage[caption=false]{subfig}% Support for small, `sub' figures and tables
\usepackage[natbibapa,nodoi]{apacite}
\usepackage{graphicx}
\usepackage{natbib}
\usepackage{soul}
\usepackage{url}
\usepackage{eurosym}
\usepackage{listings}
\usepackage{xcolor}
\usepackage{booktabs}
\usepackage{lscape}

\usepackage{pifont} 
\newcommand{\cmark}{\ding{51}}
\newcommand{\xmark}{\ding{55}}

\colorlet{punct}{red!60!black}
\definecolor{background}{HTML}{FFFFFF}
\definecolor{delim}{RGB}{20,105,176}
\colorlet{numb}{magenta!60!black}

\lstdefinelanguage{json}{
    basicstyle=\small\ttfamily,
    showstringspaces=false,
    breaklines=false,
    tabsize=4,
    backgroundcolor=\color{background},
    literate=
     *{0}{{{\color{numb}0}}}{1}
      {1}{{{\color{numb}1}}}{1}
      {3}{{{\color{numb}3}}}{1}
      {4}{{{\color{numb}4}}}{1}
      {5}{{{\color{numb}5}}}{1}
      {6}{{{\color{numb}6}}}{1}
      {7}{{{\color{numb}7}}}{1}
      {8}{{{\color{numb}8}}}{1}
      {9}{{{\color{numb}9}}}{1}
      {:}{{{\color{punct}{:}}}}{1}
      {,}{{{\color{punct}{,}}}}{1}
      {\{}{{{\color{delim}{\{}}}}{1}
      {\}}{{{\color{delim}{\}}}}}{1}
      {[}{{{\color{delim}{[}}}}{1}
      {]}{{{\color{delim}{]}}}}{1},
}

\lstdefinelanguage{json2}{
    basicstyle=\small\ttfamily,
    showstringspaces=false,
    breaklines=true,
    tabsize=3,
    frame=lines,
    backgroundcolor=\color{background},
    literate=
      {:}{{{\color{punct}{:}}}}{1}
      {,}{{{\color{punct}{,}}}}{1}
      {\{}{{{\color{delim}{\{}}}}{1}
      {\}}{{{\color{delim}{\}}}}}{1}
      {[}{{{\color{delim}{[}}}}{1}
      {]}{{{\color{delim}{]}}}}{1},
}

\theoremstyle{plain}% Theorem-like structures provided by amsthm.sty

\theoremstyle{definition}

\theoremstyle{remark}

\begin{document}

\articletype{ARTICLE TEMPLATE}% Specify the article type or omit as appropriate

\title{Efficient micro data centers deployment for mobile healthcare monitoring systems in IoT urban scenarios}

\author{
\name{Kevin Henares\textsuperscript{a}, José L. Risco-Martín\textsuperscript{a}\thanks{CONTACT José L. Risco-Martín. Email: jlrisco@ucm.es}, José L. Ayala\textsuperscript{a} and Román Hermida\textsuperscript{a}}
\affil{\textsuperscript{a}Dept. of Computer Architecture and Automation, 
Complutense University of Madrid, Calle Prof. José García Santesmases, 9, 28040 Madrid, Spain}
}

\maketitle

% max: 200 words (https://www.tandfonline.com/action/authorSubmission?show=instructions&journalCode=tjsm20&utm_source=CPB&utm_medium=cms&utm_campaign=JPD14564&#prep)
% Internet of Things, Modeling and Simulation Based Systems Engineering, edge and fog computing, discrete-event system specification, clustering

\begin{abstract}
In the last decade, the Internet of Things paradigm has caused an exponential increase in the number of connected devices. This trend brings the Internet closer to everyday activities and enables data collection that can be used to create and improve a great variety of services and applications. Despite its great benefits, this paradigm also comes with several challenges. More powerful storage and processing capabilities are required to service all these devices. Additionally, the need to deploy and manage the infrastructure to efficiently support these resources continues to pose a challenge. Modeling and simulation can help to design and analyze these scenarios, providing flexible and powerful mechanisms to study and compare different strategies and infrastructures. In this scenario, Micro Data Centers (MDCs) can be used as an effective way of reducing overwhelmed Cloud Data Center infrastructures. This paper explores an M\&S methodology to study the overall power consumption of a healthcare IoT scenario. The patients wear non-intrusive monitoring devices that periodically generate tasks to be executed in MDCs. We extract the layout of existing  urban infrastructures, simulate the monitored population's behavior, and compare the power consumption of several data center configurations.
\end{abstract}


\section{Introduction}
\label{sec:intro}

% Data generation
The data generation ratio worldwide is increasing exponentially over the years. In every scope, the domain-specific information is progressively taken into account to create valuable knowledge that can help us understand our reality and make our procedures and technologies more effective and efficient. According to the International Data Corporation (IDC), worldwide created, captured, and replicated data will grow to 175 zettabytes by 2025. Since 2018 (in which 33 zettabytes were registered), this represents an increment of 530\% and a compound annual growth rate of 61\% \citep{reinsel2018data}.

This increment has been favored by the growing number of connected devices and the development of new and improved IoT ecosystems. IDC predicts that by 2025 there will be 55.7 billion connected devices worldwide, 75\% of which will be connected to an IoT platform. They estimate data generated from connected IoT devices to be 73.1 ZB by 2025, growing from 18.3 ZB in 2019~\citep{idc2020iot}. Moreover, this growth in the total number of IoT devices is projected to provide substantial economic and social benefits in the way of cost savings, value creation, and productivity improvements~\citep{castillo2015projecting}. Improved industrial monitoring and automation techniques will help to minimize failures, reduce waste, and optimize processes. Smart cities are expected to contribute to this trend, paving the way for the introduction of intelligent and unmanned transportation~\citep{tokody2017overview} and optimized deliveries~\citep{wang2019demystifying}, and optimizing their  infrastructures~\citep{barba2012smart}~\citep{masera2018smart}~\citep{dey2000context}.

In healthcare, accurate patient monitoring and pharmaceutical management, added to predicting  risk factors in highly-impact diseases, will result in substantial cost savings. These predictions will be facilitated by the implementation of Wireless Sensor Networks (WSNs), where patients with specific target diseases wear unintrusive wearable devices that allows to continuously monitor patients state, registering variables like heart rate, Electrocardiogram (ECG), electrodermal activity, Electroencephalography (EEG), or oxygen saturation. These variables, related to the autonomic nervous system (ANS), can be modeled for associating them with specific disease symptoms or outcomes, generating useful predictions for patient diagnosis and treatment~\citep{pagan2015robust, ali2020smart}. 

These scenarios come accompanied with a need for storage infrastructures and computing capabilities. This trend has led to the increasing use of data centers to store and process data traditionally located at endpoints. Cloud computing is one of the enabling
platforms to support these needs. The National Institute of Standards and Technology (NIST)~\citep{mell2011nist} defines cloud computing as "\textit{...a model for enabling ubiquitous, convenient, on-demand network access to a shared pool of configurable computing resources (e.g., networks, servers, storage, applications, and services) that can be rapidly provisioned and released with minimal management effort or service provider interaction}". Armbrust et al.~\citep{fox2009above} simplify this concept defining \textit{cloud} as the “\textit{datacenter hardware and software that provide services}". However, in addition to raw computing and storage, cloud computing providers usually offer multiple software services, APIs, and development tools that allow developers to build seamlessly scalable applications upon their services~\citep{voorsluys2011introduction}. This Infrastructure as a Service (IaaS) model approaches large and powerful cloud infrastructures to individuals, allowing them to develop solutions to simplify and optimize different aspects of everyday life through a new range of innovative services and applications. Through its regular use, it is possible to improve the systems' performance, easily make use of these storing and processing capabilities, and reduce systems overall cost.

However, there are also some drawbacks when using these services. Some applications may not assume the latency derived from the intermediate communications, or may be restricted by security or privacy concerns. In these cases, Fog Computing is often used. This paradigm tries to solve these aspects by making available these infrastructures to end-users that provides several advantages. In addition to the lower latencies derived from the tightly coupled infrastructures, it enables the use of mid-range IoT protocols and reduces problems related to bandwidth bottlenecks. Also, the reliability of the connection is increased due to the existence of multiple interconnected channels. Some fog nodes include industrial controllers, Micro Datacenters (MDCs), and video surveillance cameras.

All these technologies enable the development of connected healthcare scenarios, where networked sensors are being placed either in the body or in the living environments of patients to continuously extract different types of data. These data, combining several technologies like Cloud Computing, Big Data, the Internet of Things, and Machine Learning, are being aggregated and processed to create valuable and meaningful insights regarding the patients' lifestyle, habits, and health conditions. In this way, these proactive systems are helping to evolve the traditional concept of medicine, favoring the prognosis over the classical post-facto reactive paradigm ~\citep{saha2017health}. However, while incorporating all these technologies, healthcare scenarios are becoming larger and more complex. They often include many heterogeneous and dynamic components, that constantly evolve to incorporate new features over time. Modeling and simulation (M\&S) is progressively gaining acceptance in the healthcare field for studying and tackling such complexity. M\&S shows excellent potential for comparing methodologies and scenarios, visualizing workflows, or even monitoring and increasing the performance of health care procedures through their integration with information system applications~\citep{gaba2007future}. Moreover, M\&S can be used to simulate large amounts of configurations in manageable times, which would be too expensive or infeasible to execute in real-world settings~\citep{erdemir2020credible}.

This paper uses these benefits to perform an M\&S-driven analysis of the energetic impact of MDCs geographical location in an urban healthcare scenario.
%This paper analyzes the impact of the geographical location of MDCs in a healthcare scenario. 
This scenario includes a population of migraine patients, being continuously monitored by non-intrusive and portable devices. These devices periodically send information to MDCs to retrieve predictions generated with the corresponding patient-oriented models. These predictions can help migraine patients avoiding the pain by taking measures before the pain starts.

Migraine is one of the most disabling neurological diseases. It affects around 10\% of population worldwide~\citep{linde2012cost} and 15\% in Europe~\citep{stovner2010prevalence}. 
Apart from the pain phase, migraine disease includes other less-known symptoms. Premonitory or prodromic symptoms may occur from three days to hours before the pain starts~\citep{giffin2003premonitory}. They are subjective, varied, and include changes in mood, appetite, sleep, etc. Auras occur in one-third of the cases~\citep{headache2004international} and appear within 30 minutes before the onset of pain. It consists of a short period of visual disturbance. Postdrome are symptoms that occur after the headache. Some of the most common are tiredness, head pain, or cognitive difficulties. They are present in 68\% of the patients, and they have an average duration of 25.2 hours~\citep{kelman2006postdrome}.
Moreover, migraine sufferers are more prone to suffer from other diseases such as fatigue, anxiety, or cardiovascular problems, which leads to high costs for private and national health systems. In Europe, it is estimated that migraine leads to direct and indirect costs of \euro{1,222} per patient per year~\citep{linde2012cost}. 
It is difficult to estimate the onset of pain to make an adequate intake of drugs. The time response of the drugs' pharmacokinetics (the mechanisms of absorption and distribution of substances in an organism) does not match the long times of the vague predictive symptoms, or the short times of the urgent auras. So, most migraine sufferers wait until the onset of pain to take the rescue medication. The delayed intake reduces the effectiveness of the treatment. 

In the remainder of this article, Section~\ref{sec:background} reviews existing work related to healthcare monitoring simulation scenarios, and Cloud/Fog-related analysis and processing methodologies and platforms. Section~\ref{sec:methodology} gives further details about the use case and the goals covered by this article. The results are then discussed in Section~\ref{sec:results}. Finally, Section~\ref{sec:conclusions} present the conclusions and provide some suggestions for future work.

%%%%%%%%%%%%%%%%%%%%%%%%%%%%%%%%%%%%%%%%%%%%%%%%%%%%%%%%%%%%%%%%%%%%%%%%%%
%%%%%%%%%%%%%%%%%%%%%%%%%%%%%%%%%%%%%%%%%%%%%%%%%%%%%%%%%%%%%%%%%%%%%%%%%%

\section{State of the art}
\label{sec:background}

When designing and analyzing data-driven IoT scenarios, multiple factors must be considered. Aspects such as the application's placement, the system's architecture, the storing and processing resources, the communication networks and protocols, the job allocation policies, or the energy constraints, may have critical importance in the full service's usability and performance. To ease the development of such systems, several simulation frameworks have been developed over the years. Section~\ref{sec:background_frameworks} describes some of the most popular IoT simulation frameworks. Section~\ref{sec:background_related_work} discusses several principles and approaches chosen by other authors when developing this type of scenario. 

\subsection{M\&S for the development of complex systems}
Complex systems are usually understood as organizations including a large number of interacting components. They often have a heterogeneous and dynamic nature, and progressively increase their complexity over time. Some examples of complex systems are communication infrastructures, the Earth climate system, biologic organisms, or socio-economic organizations as cities. Mathematical models are often used as an effective tool to study and tackle their inner complexity, providing precise mechanisms to represent real problem situations and helping to make decisions faster and more accurately. Furthermore, when analytical solutions are not enough to study such systems, simulation models can describe systems' behavior. These computerized models extend these analytical properties through the use of simulation methodologies or formalisms (e.g., Cellular Automata, Machine Learning, fuzzy models) and algorithms reproducing their life-cycle based on these descriptions.

The development of such simulation models starts with analysis over real-life phenomena to extract different types of knowledge (i.e., mathematical properties, behavioral rules) and create a conceptual model. Simulation models derive from this specification, being implemented in specific programming languages following the guidelines described by the selected M\&S framework. Through its execution with a simulation engine and varying the input parameters and configurations, it is possible to obtain a wealth of information that helps modelers understand reality and create more accurate specifications.
Over time, models increase their complexity in many ways. They accumulate knowledge of several disciplines and increase the level of detail of certain phenomena representations as the development iterations are completed. The same model can combine low-level and high-level descriptions and analyze a problem with different levels of detail. Moreover, the combination of the size and diversity of these models often leads to large-scale behaviors known as emergent behaviors. Therefore, as the systems grow, the properties of the entire system become very different from those
of its components. Although several authors have reviewed the theoretical foundation for modeling emergent behavior in the literature and several tools have been created to facilitate the specification of these phenomena, reproducing emergent behaviors in artificial environments is still a challenging task \citep{Mittal2018}. As the models produced in the modeling workflow represent simplified versions of the existing systems, the inner omission of some low-level details implies an information loss that may cancel these emerging behaviors. To avoid this, the model has to be supported by solid hypotheses or theoretical constructs and be extensively simulated to check its correctness. 
The use of M\&S formalisms contributes to this objective, offering clear, reusable, and unambiguous specifications that deal with the complexity and particularities of these systems. Some examples, matured in academia for decades, are Petri Nets, Markov Chains, Cellular Automata, or Discrete Event System Specification (DEVS).

\subsection{IoT simulation frameworks}
\label{sec:background_frameworks}

Some simulators allowing the specification of IoT scenarios are:

\begin{itemize}
    \item \textbf{CloudSim}~\citep{calheiros2011cloudsim} is an open-source simulation framework developed at the Cloud Computing and Distributed Systems (CLOUDS) laboratory of the University of Melbourne. It allows modeling, simulation, and experimentation of Cloud computing infrastructures and application services. It also includes an end-to-end Cloud network architecture that utilizes BRITE topology~\citep{medina2001brite} for modeling link bandwidth and associated latencies. This framework has been used as a basis to develop a multitude of specialized frameworks.
    %\item \textbf{EdgeCloudSim}~\citep{sonmez2018edgecloudsim} extends the functionality offered by CloudSim so that it can be efficiently used for Edge Computing scenarios. Through a modular architecture, it provides support for a variety of functionality such as network modeling specific to WLAN and WAN, device mobility model, realistic and tunable load generator. It also supports the definition of scenarios with different layers of Edge servers, properly coordinated with different cloud resources. Other platforms, like \textbf{IoTSim}~\citep{zeng2017iotsim} or \textbf{MRCloudSim}~\citep{jung2012mr} focus on extending the original CloudSim implementation for enabling the simulation of IoT Big Data processing. For this purpose, they add support for the MapReduce distributed computing model. 
    
    \item \textbf{EdgeCloudSim}~\citep{sonmez2018edgecloudsim} is an edge-oriented simulator developed at the Department of Computer Engineering at the Bogazici University. It extends the functionality of CloudSim so that it can be used for Edge Computing scenarios. Through a modular architecture, it provides support for a variety of functionality such as network modeling specific to WLAN and WAN, device mobility model, realistic and tunable load generator. It also supports the definition of scenarios with several Edge server layers, properly coordinated with different cloud resources. To facilitate these definitions, it also includes orchestration modules to model the organization of the different types of resources. 
    
    \item \textbf{EmuFog}~\citep{mayer2017emufog} is an an extensible emulation framework for the definition of Fog Computing scenarios. It allows the definition of fog infrastructures and emulates real applications and workloads by embedding Docker images in the scenario nodes. All components of EmuFog are extensible and replaceable by custom-built components designed for specific scenarios. Also, it allows loading the designs performed in network topology generators as BRITE~\citep{medina2001brite}, facilitating the import of real-world topology datasets.

    \item \textbf{FogNetSim++}~\citep{qayyum2018fognetsim++} is a toolkit for the  modeling and simulation of distributed fog environments. It is built on the top of OMNeT++, a discrete-event simulator oriented to the development of network simulators. It allows us to incorporate customized mobility models and fog node scheduling algorithms, and manage handover mechanisms. It bases the construction of IoT scenarios on three main types of modules: (i) end devices, (ii) fog nodes, and (iii) brokers. In these scenarios, the end devices contain the actual sensors and generate the processing requests to upper layers. The broker receives these requests and sends them to the suitable fog nodes. The broker nodes are also connected to a backbone network which connects them to cloud data centers. FogNetSim++ allows researchers to incorporate their request scheduling and handover algorithms, simplifying the study of the energetic impact of different delivery strategies.
    
    \item \textbf{FogTorch}~\citep{brogi2017fogtorch} is a Java tool for the definition of QoS-aware IoT applications to Fog infrastructures. It divides the definition of the scenario into three levels: (i) IoT devices, (ii) one or more layers of Fog computing nodes, and (iii) one or more cloud data centers. The Cloud concept is simplified in the FogTorch simulation model, understood as a virtually unlimited amount of hardware capabilities. This limitation constrains the scenario definition, but eliminates the need to describe particular cloud topologies and infrastructures and simplifies the definition of any SaaS, PaaS, or IaaS service. FogTorch allows specifying different Quality of Service (QoS) profiles based on pairs of latency and bandwidth values, associating them with specific network links. 
    
    \item \textbf{GloudSim}~\citep{di2015gloudsim} is a distributed cloud simulator that aims to reproduce the Google cloud environment, allowing to define its cluster infrastructures and simulate different types of events. It allows defining dynamic resource consumption and priority levels for the emulated jobs and reproducing kill/evict events. It is compatible with the Google traces format and includes several interfaces and scripts to extract the information from these CSV trace files. Among other parameters, it allows us to obtain the CPU and memory consumption, the number of simultaneous active jobs, and the workload processing ratio. GloudSim has been published under the GNU GPL v3 license. 
    
    \item \textbf{iCanCloud}~\citep{nunez2012icancloud} is a simulation platform aimed to model and simulate cloud computing systems. The main objective of iCanCloud is to predict the trade-offs between cost and performance of a given set of applications executed in specific hardware, and then provide users helpful information about such costs. It allows modeling the cloud architecture, includes a hypervisor module for managing and comparing different brokering policies, and enables the extraction of detailed energy consumption information for each hardware component of the whole infrastructure. It is also possible to specify custom policies to analyze the impact of energy consumption on the overall system performance, facilitating the study of trade-offs between performance and energy consumption.
    
    \item \textbf{iFogSim}~\citep{gupta2017ifogsim} is a toolkit for modeling and simulation of resource management techniques in the Internet of Things, derived from the CloudSim framework. It allows the study of different resource management policies applicable to fog environments concerning their impact on latency (timeliness), energy consumption, network congestion, and operational costs. It simulates edge devices, cloud data centers, and network links to measure performance metrics. One of its main contributions is the \textit{Sense-Process-Actuate}, which allows defining scenarios where sensors publish data  periodically or based on events, and devices in the fog layer subscribe to these data streams. %In  obtaining different insights and causing responses in actuators modules.
    
    \item \textbf{IoTSim}~\citep{zeng2017iotsim} is a CloudSim-based IoT simulator that allows to specify and execute IoT big data scenarios. It organizes this specification in six layers: (i) the Core Simulation Engine Layer provides core functionalities as the creation of cloud elements, communication among components, and management of the simulation time, (ii) the CloudSim Simulation Layer provides support for modeling and simulation of Cloud-based simulation environments, (iii) the Storage Layer includes different storage types as Amazon S3, Azure Blob Storage, and HDFS, (iv) the Big Data Processing Layer the processing of the data generated by the IoT devices through Map-Reduce or a streaming computing model, and (v) the User Code Layer includes several utilities to facilitate the definition and validation of IoT scenarios.

    \item \textbf{Mercury}~\citep{cardenas2020mercury} is a Modeling, Simulation, and Optimization framework to analyze the dimensioning and the dynamic operation of real-time fog computing scenarios. It has been developed by the Integrated Systems Laboratory at the Technical University of Madrid, upon the Python xDEVS API. It allows to specify 2D Mobility scenarios and includes a 5G-based model. It organizes the scenario definition in six layers: (i) IoT devices layer, (ii) edge federation layer, (iii) access points (AP) layer, (iv) radio interface layer, (v) core network layer, and (vi) Crosshaul layer. It also includes utilities to easily selecting the APs and Micro Data Centers (MDCs) optimal location and generating different output plots to study the results of the simulations. 
    
    \item \textbf{SFIDE}~\citep{penas2017sfide} is a simulation framework developed by the Embedded Systems Laboratory at the École Polytechnique Fédérale de Lausanne (EPFL). It allows the customization of the data centers architecture, including both the servers arrangement in the room, the cooling equipment, and the workloads' characterization to be executed. The real benefit of SFIDE resides in its capability to implement, test, and assess arbitrary workload allocation strategies and cooling control policies. 
    
    \item \textbf{YAFS} (Yet Another Fog Simulator)~\citep{lera2019yafs} is a fog computing simulator developed at the Department of Mathematics and Computer Science of the University of the Balearic Islands (Spain). It is implemented through Simpy, a simulator for the generation of discrete-event scenarios. It focuses on the analysis of applications design and deployment through the use of customized and dynamic strategies. YAFS allows representing the relationships among applications, network connections, and infrastructure elements, enabling the integration of application modules, workload location strategies, and path routing and scheduling. The resulting computational and transmission results can be exported as CSV files.
\end{itemize}

Table~\ref{tbl:fog_simulators} summarizes these IoT simulators, showing some of their features and capabilities. After the simulator name column, it shows the programming language in which they are implemented, and its support for defining Edge, Fog, and Cloud computing infrastructures. The next columns show if they include a graphical GUI to graphically design the scenarios, if they support Map-Reduce computing strategies, and if they allow defining the network components and links. Finally, the last two columns indicate if the simulators allow extracting raw data or visualizations related to the power consumption and network communication latency, respectively. Among these simulators, it is worth noting that only a few base their development on simulation formalisms. Among these, we found Mercury and SFIDE, which are built upon the xDEVS framework, and YAFS, which uses SimPy for its construction. A formal simulation basis offers many advantages in developing simulators and systems, providing unambiguous specifications that help manage the complexity of large systems and the interaction among all their heterogeneous components. Moreover, they often offer mechanisms to separate the model definition from the simulation itself. In this way, they facilitate the scalability of simulators' maintainability while favoring more robust and reliable developments.

In this paper, we use the Mercury framework to analyze the power consumption of data centers receiving processing requests of IoT devices in a Wireless Sensor Body Network (WSBN). We selected this framework since it includes the IoT layers we need to incorporate and avoids complexities related to the cloud layer and network infrastructures, not needed in this research. Additionally, Mercury is the only one that provides support for 2D mobility, facilitating our development. With this framework, we model a Micro Data Centers (MDCs) architecture and components, as well as the services related to the data processing task requested by the agents. Also, it provides mechanisms to estimate the optimal locations of the data centers based on the patients' movement over the scenario, and includes mechanisms to export and visualize the simulations' results.

\begin{landscape}
% \usepackage{multirow}
\begin{table}[]
\caption{Comparison of Fog/Cloud simulators for defining IoT applications and scenarios.}
\label{tbl:fog_simulators}
\centering
\begin{tabular}{@{}cccccccccc@{}}
\toprule
Simulator             & \begin{tabular}[c]{@{}l@{}}Programming\\ Language\end{tabular} & Edge  & Fog  & Cloud & GUI  & Map-Reduce & \begin{tabular}[c]{@{}c@{}}Network\\ configuration\end{tabular} & \begin{tabular}[c]{@{}c@{}}Power\\ consumption \\ analysis\end{tabular} & \begin{tabular}[c]{@{}c@{}}Communication\\ latency\\ analysis\end{tabular} \\
\midrule
CloudSim     & Java                                                           & \xmark  & \xmark & \cmark  & \xmark & \xmark       & \xmark                                                            & \cmark                                                                    & \cmark                                                                      \\
CloudExp     & Java                                                           & \cmark & \xmark & \cmark  & \cmark & \cmark       & \cmark                                                            & \cmark                                                                    & \cmark                                                                      \\
EdgeCloudSim & Java                                                           & \cmark  & \xmark & \cmark  & \xmark & \xmark       & \cmark                                                           & \cmark                                                                    & \cmark                                                                      \\
EmuFog                & Python                                                         & \cmark  & \cmark & \xmark  & \xmark & \xmark       & \cmark                                                            & \xmark                                                                    & \cmark                                                                       \\
FogNetSim++  & C++                                                            & \cmark  & \cmark & \cmark  & \cmark & \xmark       & \cmark                                                            & \cmark                                                                    & \cmark                                                                       \\
FogTorch              & Java                                                           & \cmark  & \cmark & \cmark  & \xmark & \xmark       & \xmark                                                            & \xmark                                                                    & \cmark                                                                       \\
GloudSim     & Java                                                           & \xmark  & \xmark & \cmark  & \xmark & \xmark       & \cmark                                                            & \xmark                                                                    & \xmark                                                                       \\
iCanCloud    & C++                                                            & \xmark  & \xmark & \cmark  & \cmark & \xmark       & \cmark                                                            & \xmark                                                                    & \xmark                                                                       \\
iFogSim      & Java                                                           & \xmark  & \cmark & \cmark  & \cmark & \xmark       & \cmark                                                           & \cmark                                                                    & \cmark                                                                      \\
IOTSim       & Java                                                           & \xmark  & \xmark & \cmark  & \xmark & \cmark       & \cmark                                                            & \xmark                                                                    & \cmark                                                                       \\
Mercury      & Python                                                         & \cmark  & \cmark & \xmark  & \xmark & \xmark       & \cmark                                                            & \cmark                                                                    & \cmark                                                                       \\
SFIDE         & Java                                                         & \cmark  & \cmark & \cmark  & \xmark & \xmark       & \cmark                                                            & \cmark                                                                    & \cmark  \\
YAFS         & Python                                                         & \cmark  & \cmark & \cmark  & \xmark & \xmark       & \cmark                                                            & \cmark                                                                    & \cmark  \\
\bottomrule
\end{tabular}
\end{table}
\end{landscape}

\subsection{Data-driven IoT scenarios: related work}
\label{sec:background_related_work}

When considering the usability and performance of an IoT scenario, several parameters may be considered. From an end-user point of view, some aspects like the perceived latency, the battery life of the nodes, and the overall cost, have a critical impact on the acceptance of IoT products. The optimization of these key aspects can be tackled when developing the IoT scenario by optimizing the power consumption, the processing workflows, and the required bandwidth. The application placement, communication protocols, or job allocation policies are some of the decisions that can alter the usability and performance of the final product. For instance, assigning more processing responsibilities to the end-nodes can highly reduce the latency perceived by the users, but significantly impact the battery life and cost of the product. On the contrary, delegating processing to cloud services can lead to substantial cost savings and higher reliability and performance, but its use implies the acceptance of higher latency and more complex system architecture. Usually, fog computing offers a good trade-off between these two approaches, and it is especially suitable when developing low latency or real-time applications. A good example is healthcare IoT systems, which are usually latency-sensitive, show low response time, and produce a large amount of data~\citep{mutlag2019enabling}. As a result, intensive use of fog computing has been made in this area. For instance, Tuli et al.~\citep{tuli2020healthfog} presented a fog-based smart healthcare system for automatic diagnosis of heart diseases using deep learning techniques. The data coming from different IoT devices is processed through a fog service for deducing the health status of patients and identify heart disease severity. Sahoo et al.~\citep{sahoo2016analyzing} designed a stochastic prediction model to foresee the future health condition of the groups of correlated patients based on their current health status. Aazam et al.~\citep{aazam2015fog} presented a resource management model for IoTs based on MDCs, covering resource prediction,  customer type based resource estimation and reservation,  and pricing estimations. Jararweh et al.~\citep{jararweh2017software} developed a software-defined based framework to allow mobile cloud computing (MCC) services to integrate different Software-Defined Systems (SDSys) in a Mobile Edge Computing context. This software-based specification of systems abstracts the complexity of large infrastructure deployments and is especially useful for content delivery networks, crowdsourcing, traffic management, or E-health, among others.

\begin{figure}
\centering
\subfloat[Pedestrian Dynamics.] {
    \includegraphics[width=0.47\textwidth]{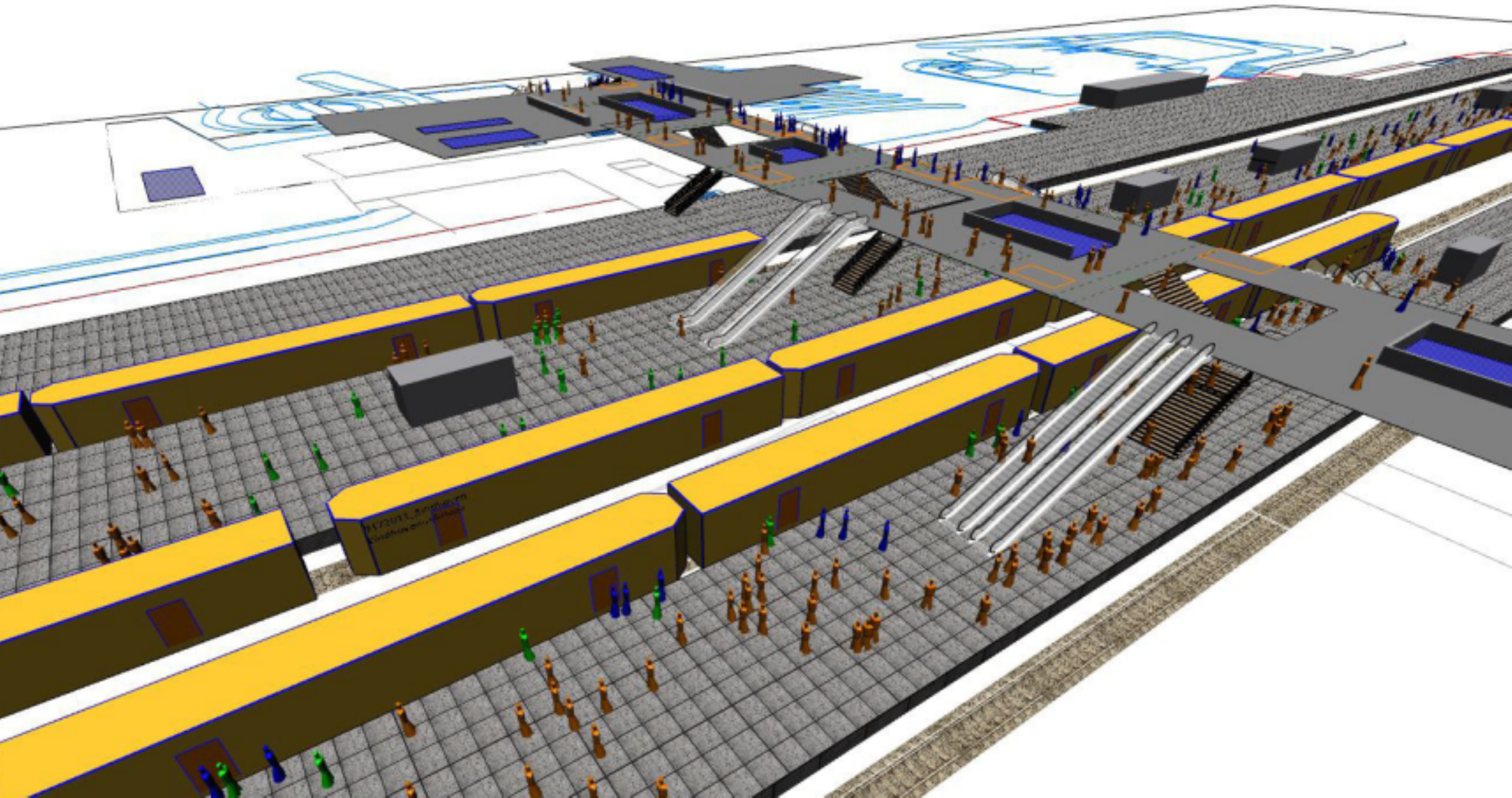}
    \label{img:int_ped_dynamics}
}
\hspace{5pt}
\subfloat[Simulation of Urban MObility (SUMO).] {
    \includegraphics[width=0.47\textwidth]{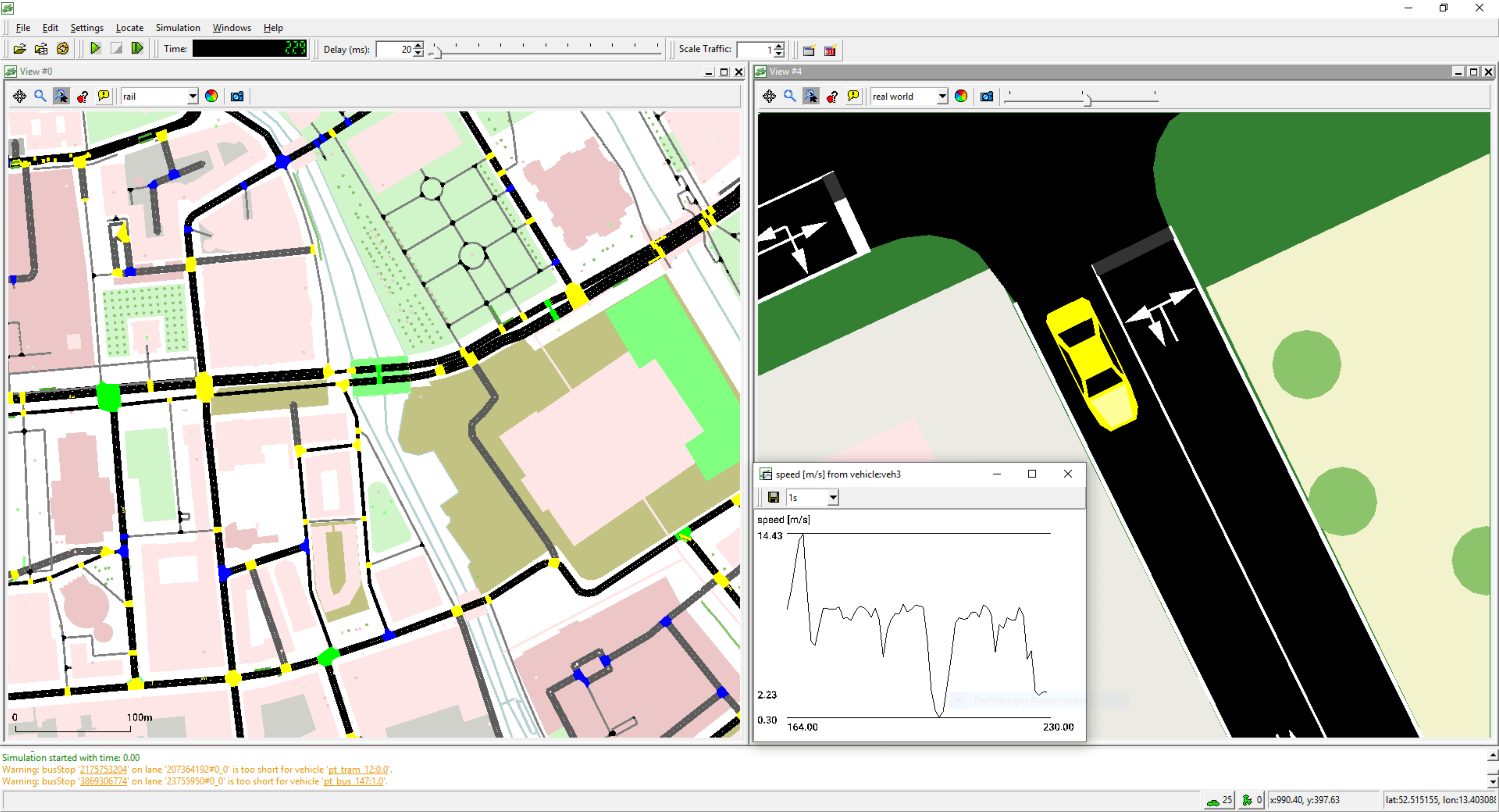}
    \label{img:int_sumo}
}

\caption{Example scenarios designed with urban simulation packages.} \label{img:int_urban_simulations}
\end{figure}

In IoT healthcare systems, one of the most important involved methods is monitoring~\citep{mutlag2019enabling}. A typical paradigm in this kind of scenario is the Modeling and Simulation (M\&S) of crowds. The representation of the behavior of a population or a group of individuals allows the development of realistic scenarios  as a basis for evaluating new technologies or methodologies. Multiple techniques are used~\citep{zhou2010crowd} to represent this behavior, such as agent-based models \citep{Zhou2012}, flow-based models \citep{Tripathi2019}, or particle system models \citep{Liu2018}. Currently, several platforms exist to simplify this crowd representation. Simulation of Urban MObility (SUMO)~\citep{behrisch2011sumo} (shown in Figure~\ref{img:int_sumo}) is an open-source traffic simulation package including  net  import  and  demand  modeling  components. It provides direct compatibility with OpenStreetMap shapefiles and allows modeling of vehicles, public transport, and pedestrians. Pedestrian Dynamics~\citep{url2020pedestrians} is a simulation environment designed to model pedestrian infrastructures or environments. It allows importing buildings geometry from multiple formats and generating 2D, 3D, and VR virtual environments. Figure~\ref{img:int_ped_dynamics} shows an scenario example developed with this software. Simulation of such large scenarios involving crowds usually requires intensive computational resources to be processed. These resources, in some cases, far exceed the capabilities of single workstations, being more suitable for its execution in clusters or data centers.

% \begin{figure}[h]
%     \centering
%     \subfigure [Pedestrian Dynamics.] {
%         \includegraphics[width=0.45\textwidth]{int_ped_dynamics}
%         \label{img:int_ped_dynamics}
%     }
%     \subfigure [Simulation of Urban MObility (SUMO).] {
%         \includegraphics[width=0.45\textwidth]{int_sumo}
%         \label{img:int_sumo}
%     }
    
%     \caption{Example scenarios designed with urban simulation packages.}
%     \label{img:int_urban_simulations}
% \end{figure}

In order to model the patients' displacement in our IoT scenario,  we have selected Pedestrian Dynamics, as it allows high configuration of the pedestrians' behavior and provides mechanisms to easily import the layout of the infrastructures. The dataset of moving pedestrians generated by this simulation platform is used as an input to Mercury's scenario to determine to which data center the patients' monitoring devices have to send their processing requests. 

%%%%%%%%%%%%%%%%%%%%%%%%%%%%%%%%%%%%%%%%%%%%%%%%%%%%%%%%%%%%%%%%%%%%%%%%%%%%%%%
%%%%%%%%%%%%%%%%%%%%%%%%%%%%%%%%%%%%%%%%%%%%%%%%%%%%%%%%%%%%%%%%%%%%%%%%%%%%%%%

\section{Experimental setup}
\label{sec:methodology}

In this paper, we consider an ambulatory monitoring system scenario where a population of migraine patients wearing health monitoring devices periodically sends some hemodynamic variables to Micro Data Centers (MDCs). These MDCs use per-patient predictive models to estimate the probability of new pain phases, generating an alert in specific patients' monitoring devices when a new pain phase is approaching. In this context, we have selected an actual urban area and modeled its infrastructures and population movement, seeking to locate the best data center locations to optimize the energy consumption of the system. 

%The main goal of this paper consist in the study of the impact that different MDCs geographical localizations have in the latency perceived by end users and the overall power consumed by the data centers.

\begin{figure}[ht]
\centering
\includegraphics[width=1\textwidth]{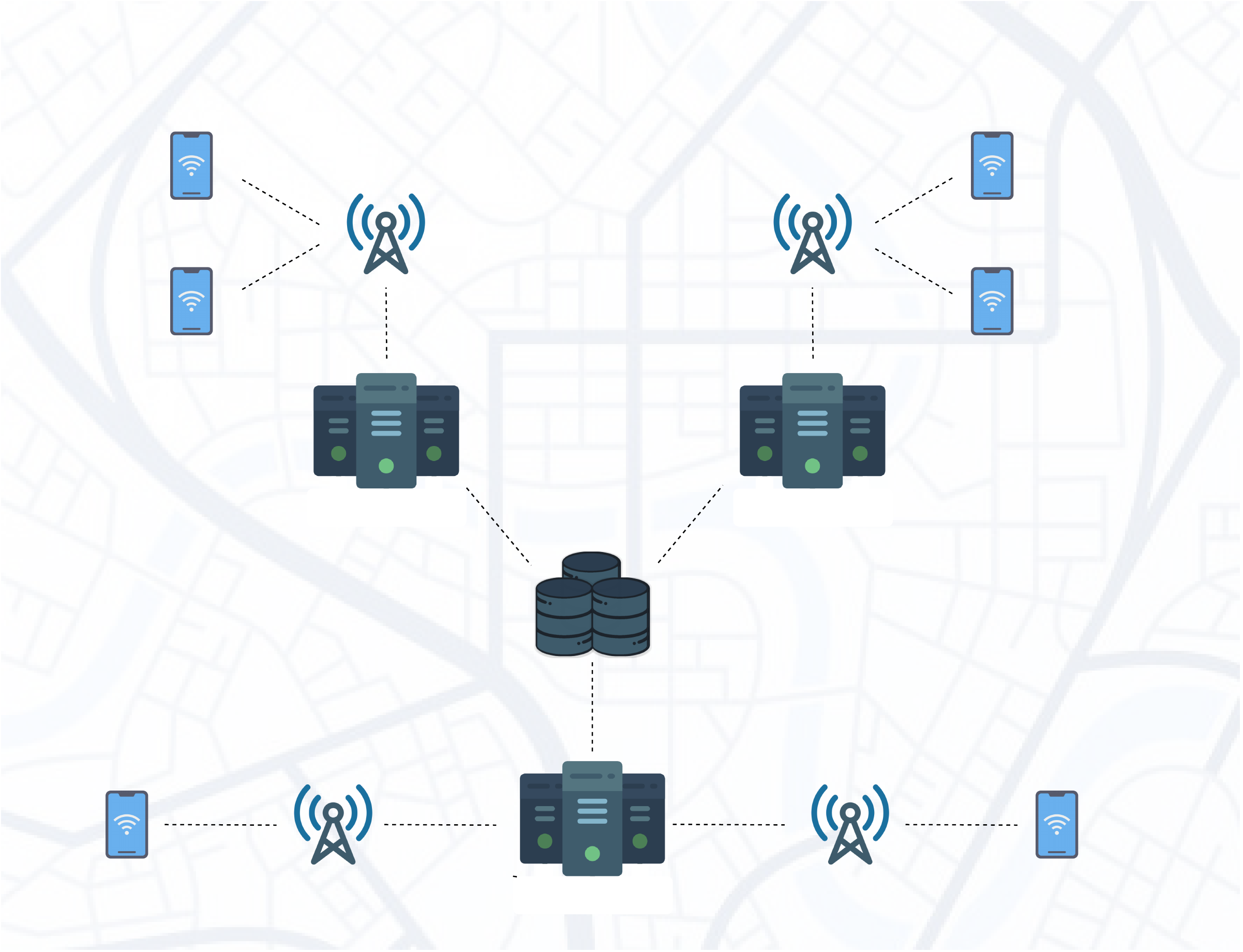}
\caption{Scenario architecture. A network of Edge Data Centers communicated with a distributed database provide connectivity to end devices through several Access Points.}
\label{img:arch_diagram}
\end{figure}

The architecture of the ambulatory monitoring scenario is represented in Figure~\ref{img:arch_diagram}. The monitoring devices carried by migraine patients periodically obtain several hemodynamic variables and send them to their smartphones. Through an app and with a predefined frequency, these data are sent periodically to the MDCs to get the predictions. This connection is performed through a network of Access Points that provide coverage to an entire metropolitan area. In this way, each phone sends the data through the nearest AP, and each AP is connected with the nearest MDC. The MDCs evaluate the data packet using per-patient custom models and return the probability of a new migraine pain phase. When this probability exceeds a threshold, the patient is alerted through the monitoring device. Moreover, there exists a shared distributed database when the different customized models are stored. Hence, when an MDC receives a prediction request corresponding to a new patient it loads the suitable model and performs the inference over it.

% Architecture and functioning of the data center
Each MDC consists of a set of racks, each one containing several servers. Sometimes, as part of the efforts to reduce the high energy consumption of data centers, idle servers are shut down until end-users request extra resources. In our simulation, for simplicity purposes, we do not use these types of strategies. This facilitates the interpretation of the results, allowing us to focus on the energy dynamically consumed by the processing units. 
%Also, the different servers have to access the per-patient predictive models before performing a model evaluation. These factors, added to the variable usage of mobile networks, produce variations in the latency perceived by the end users.
% Details of the migraine use case
For a realistic characterization of the predictive models, we set some scenario parameters based on the research conducted by Pagan et al. In their study, several types of custom migraine models have been developed, using algorithms like Subspace State Space System Identification (N4SID)~(\citep{pagan2015robust}) and Grammatical Evolution~(\citep{pagan2016grammatical}) for training the models. For the generation of these per-patient models, they perform ambulatory monitoring of the patients during a period from two weeks up to a month. The five variables implied in this process are: (i) electrodermal activity, (ii) heart rate, (iii) oxygen saturation, (iv) surface skin temperature, and (v) subjective pain level for each migraine event. This last variable is registered through an app, where the patient tracks the progress of migraine. The rest of them are registered with a portable medical monitoring device. After the model is trained in this \textit{offline} phase, it is ready to be uploaded to the servers and start generating runtime predictions of the migraines.

\begin{figure}[ht]
\centering
\includegraphics[width=1\textwidth]{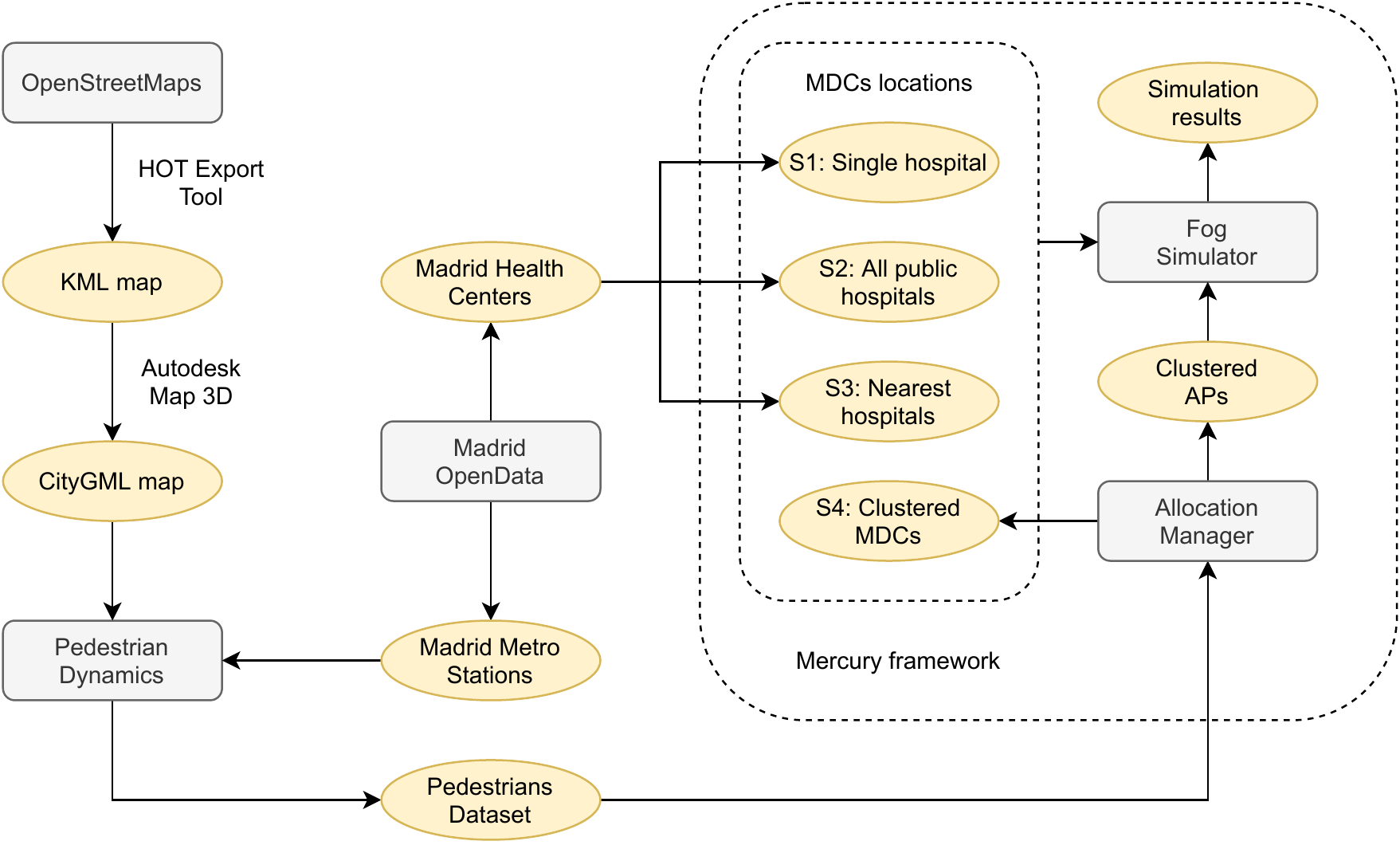}
\caption{Information workflow used in the process of modeling and simulating the pedestrians scenario.}
\label{img:pedestrians_workflow}
\end{figure}

Figure~\ref{img:pedestrians_workflow} summarizes the information workflow followed in the modeling and simulation of these scenarios. We start extracting an urban area for our scenario from OpenStreetMaps (OSM)~\citep{url2020osm}. This collaborative project gives access to detailed information of map data across much of the world, containing information such as the layout of roads and paths, buildings topology, place names, and points of interest. For these scenarios, we have selected a metropolitan area of $75km^2$ belonging to Madrid, Spain's capital city. Although OSM provides a native mechanism to extract its data, it is limited to 50000 map nodes. As this scenario corresponds to a large metropolitan area, we have used the web platform \textit{HOT Export Tool} to obtain a dataset containing the layout of all the buildings in the selected area as a KML map file. This map is then converted for compatibility reasons to CityGML~\citep{kolbe2005citygml}, an open standardized data model and exchange format to store digital 3D models of cities and landscapes, using Autodesk Map 3D. This CityGML definition is then loaded into Pedestrian Dynamics, a simulation environment designed to model pedestrian infrastructures or environments.

% \begin{figure}
% \subfigure [Map view.] {
%     \includegraphics[width=0.5\textwidth]{map_sat_with_hosp}
%     %\caption{Satellite view.}
%     \label{img:scenario_map_sat}
% }
% \subfigure[Simulation scenario.]{
%     \includegraphics[width=0.5\textwidth]{map_ped}
%     \label{img:scenario_map_ped}
% }
% \caption{Geographical area considered for the simulation. It corresponds to Madrid (Spain), has an area of around $75km$, and includes 8 hospitals.}
% \label{img:scenario_map}   
% \end{figure}

\begin{figure}
\centering
\subfloat[Map view.] {
    \includegraphics[width=0.45\textwidth]{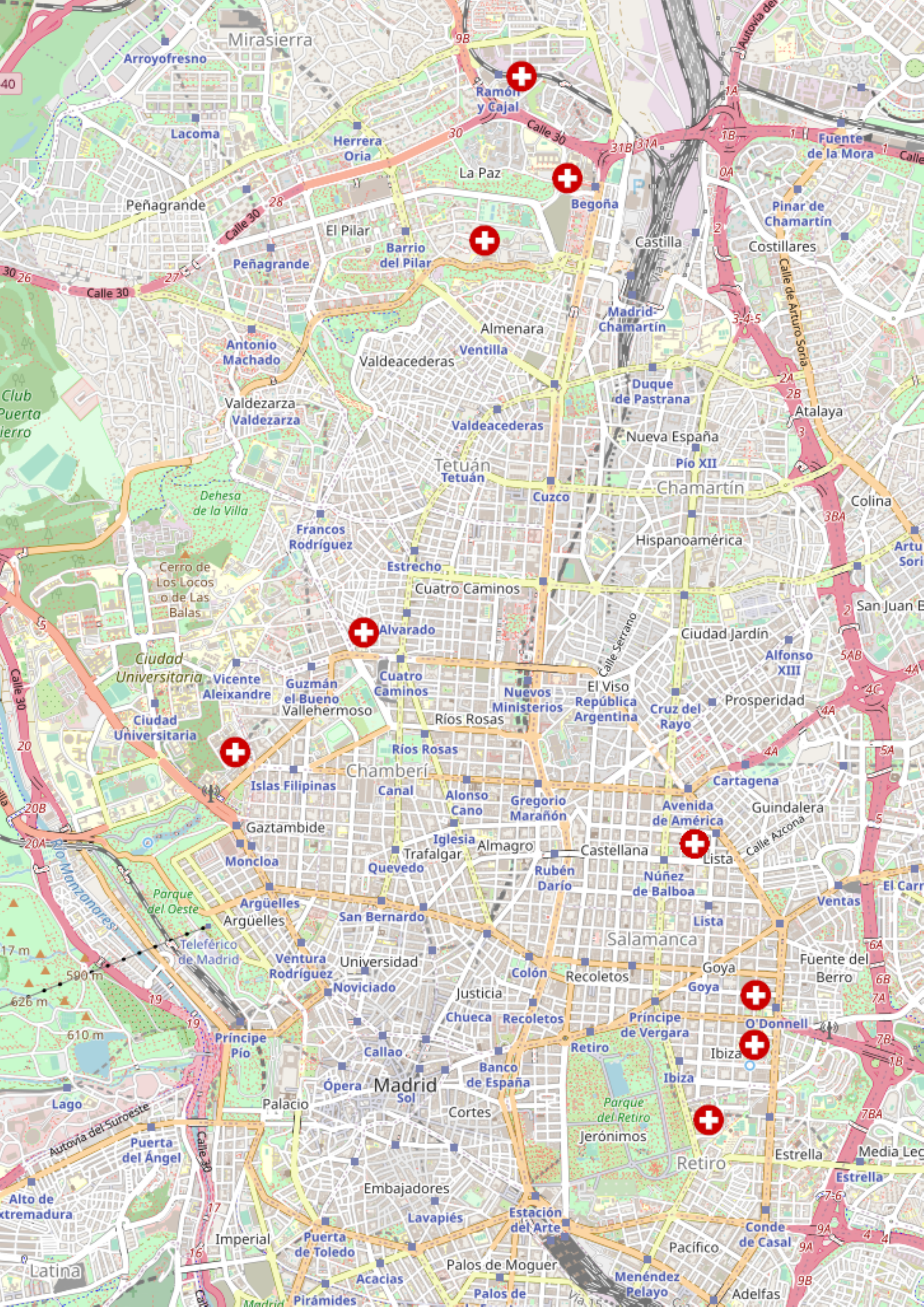}
    \label{img:scenario_map_sat}
}
%\hspace{5pt}
\subfloat[Simulation scenario.] {
    \includegraphics[width=0.45\textwidth]{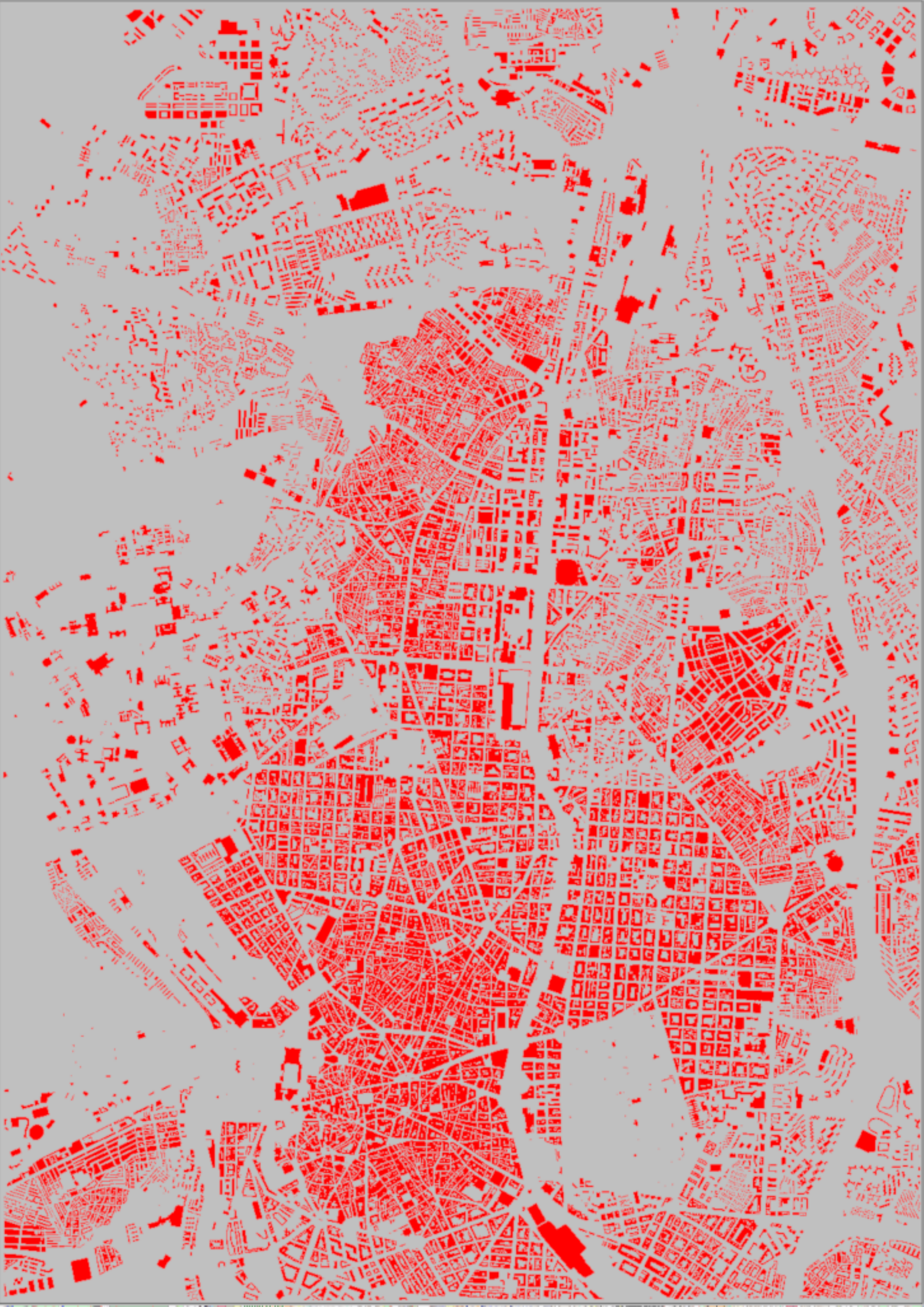}
    \label{img:scenario_map_ped}
}

\caption{Geographical area considered for the simulation. It corresponds to Madrid (Spain), has an area of around $75km^2$, and includes 9 hospitals.} \label{img:scenario_map}
\end{figure}

\begin{figure}[ht]
\centering
\includegraphics[width=0.6\textwidth]{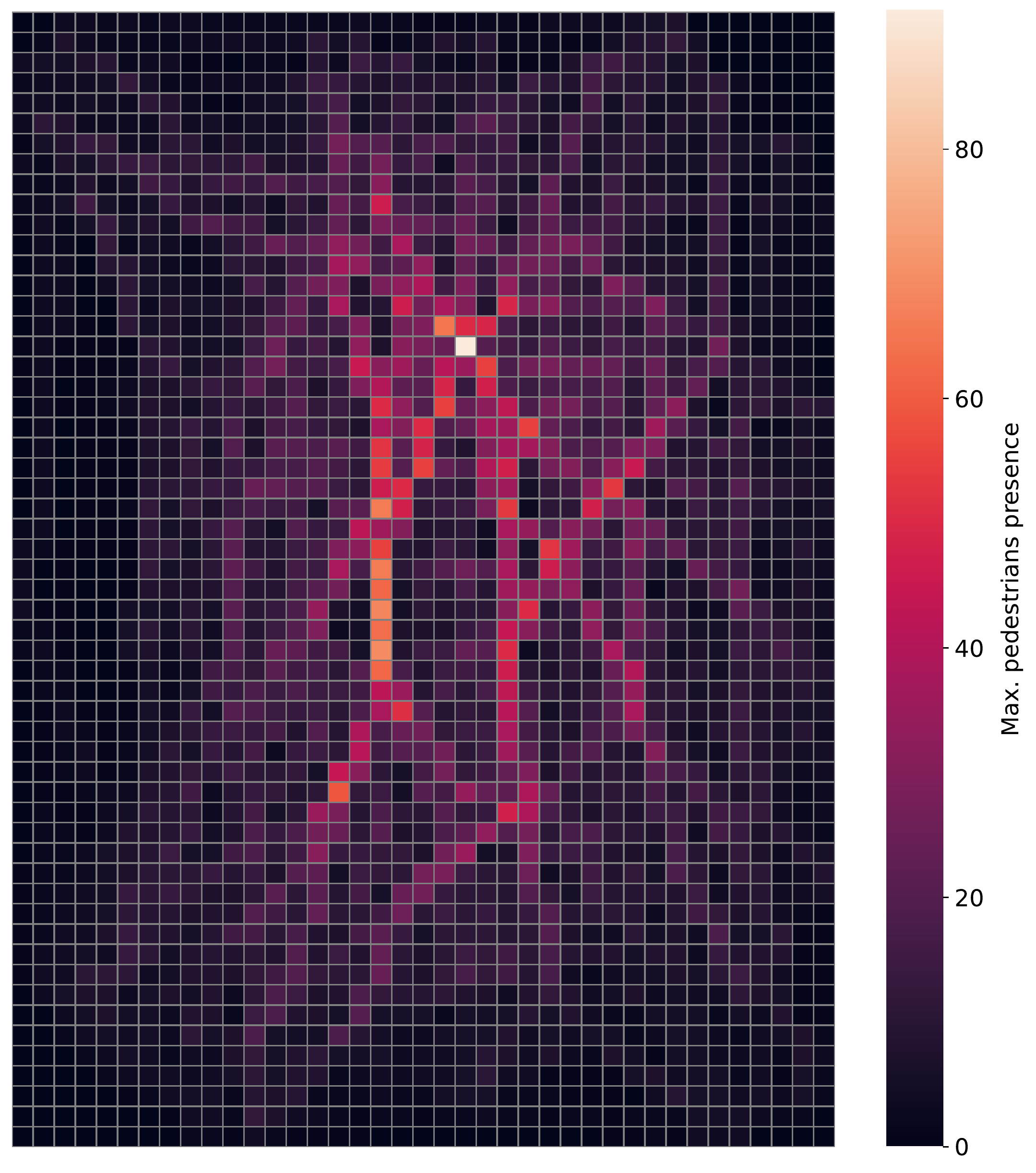}
\caption{Intermediate presence grid generated by Mercury with a time window of 60s and a grid resolution of 40. This grid is used to find the best locations of APs and MDCs through KMeans clustering.}
\label{img:kmeans_heatmap}
\end{figure}

The information of these buildings is used in the simulations as obstacles that the pedestrians have to avoid to reach their goals. The main streets and parks of the remaining space are manually marked as activity areas. These areas are used by Pedestrian Dynamics to determine the locations to which the pedestrians go during the simulation. Moreover, 139 stops of the Madrid metro network extracted from the Madrid Open Data portal were included in the map as entry/exit points. With these infrastructures already loaded into Pedestrian Dynamics, we configured the pedestrians' simulation as follows. Every 180 seconds, 200 pedestrians agents are instantiated in the entry/exit points, simulating the arrival of new trains in the metro stations according to average values recorded in the city of Madrid during peak hours \citep{AlvarezSanchez2018}. Each of these pedestrians is configured to walk until a random location included in the main streets or parks defined before, wait for a few seconds, and exit the scenario for the nearest metro stop. This simulation is performed until 10 hours of virtual time are completed (from 7 A.M. to 12 A.M. and from 4 P.M. to 9 P.M., which are the peak intervals). As a result, we obtain a dataset containing the XY coordinates of each pedestrian agent for each simulation time step, that can be used as a reference to model the behavior of the population. We have selected these parameters because they represent the most dynamic scenario: (i) the frequency of trains during the intervals selected is maximum, (ii) we have reduced the waiting time per pedestrian to a few seconds, maximizing the number of sessions opened at the MDCs, and (iii) these peak hours coincide with the maximum demand of the health system.

\begin{figure}[h]
\centering
\subfloat[Scenario 1: APs and MDCs are allocated using KMeans clustering.] {
    \includegraphics[width=0.5\textwidth]{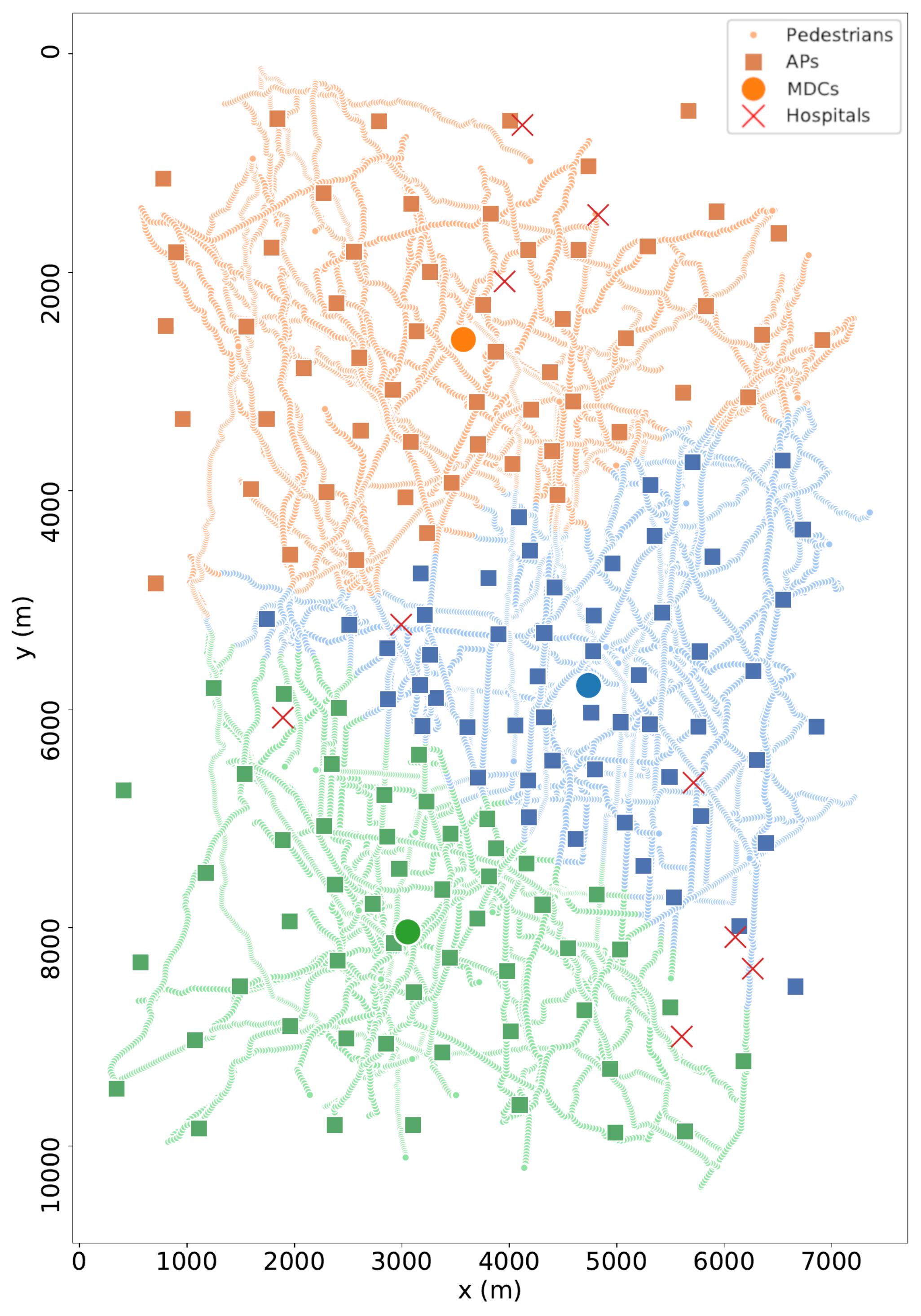}
    \label{img:edcs_aps_alloc_clus}
}
%\hspace{5pt}
\subfloat[Scenario 2: MDCs location are changed so that the correspond to the hospitals closest to the MDCs generated for Scenario 1. The same APs remain.] {
    \includegraphics[width=0.5\textwidth]{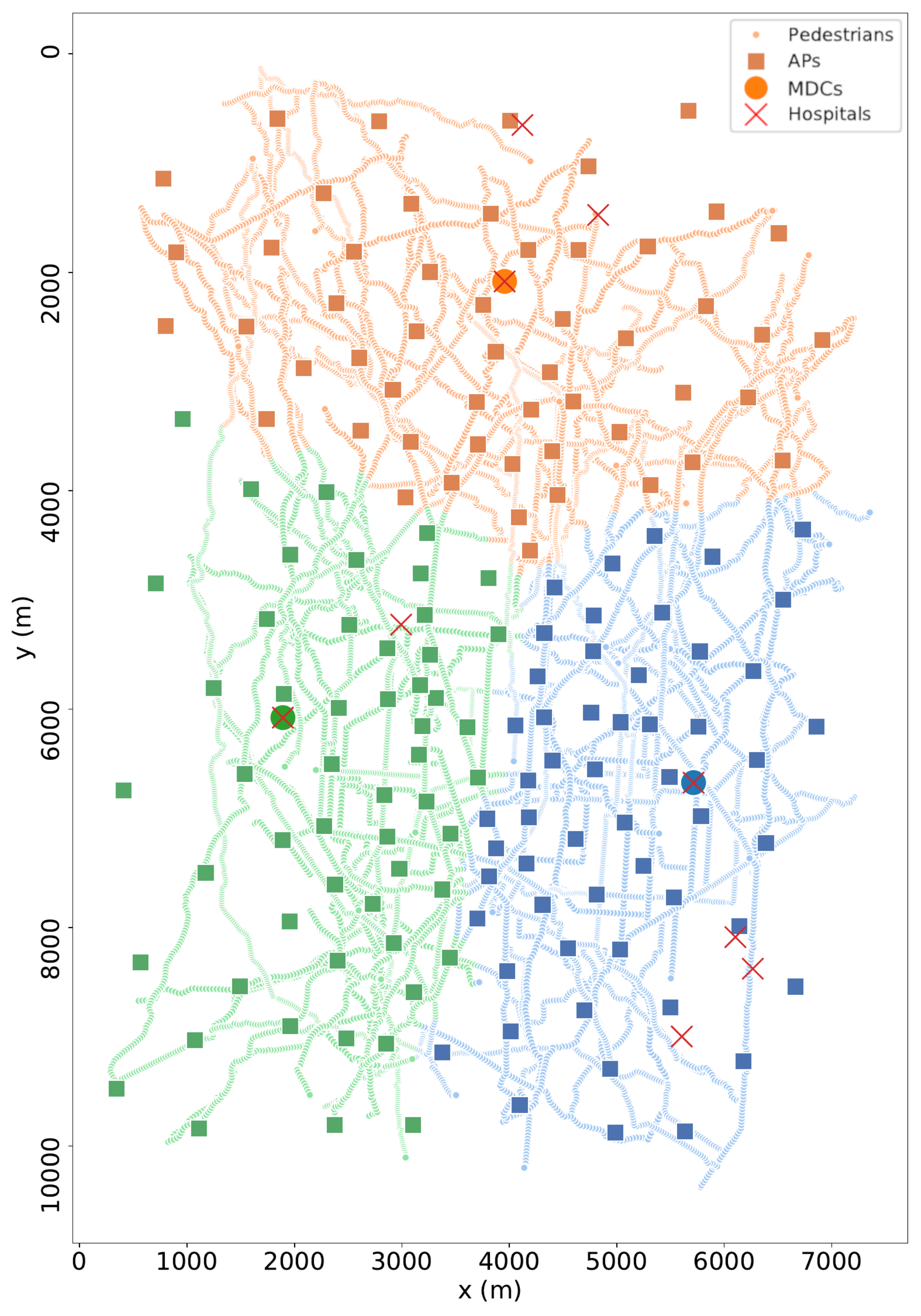}
    \label{img:edcs_aps_alloc_hosp}
}

\caption{Scenarios considered in the simulation.}
\label{img:edcs_aps_alloc}  
\end{figure}

The 10-hour dataset generated using Pedestrian Dynamics is then loaded into the Mercury DEVS framework~\citep{cardenas2020mercury}. Apart from the modules involved in the IoT scenarios simulation itself, this framework also includes some additional tools to generate datasets with optimized locations of APs and MDCs. First, this allocation method partitions the scenario map using a grid, registering the maximum presence of pedestrians in each cell considering a specific time window and grid resolution. Figure~\ref{img:kmeans_heatmap} depicts the intermediate grid. Mercury uses this information to calculate the optimized locations of APs and MDCs through a KMeans clustering process. We use this method for placing the APs location, which is common to all the scenarios. For this study, we define four alternative scenarios:
%Also, we generate a scenario with three MDCs with the locations generated with this KMean cluster as a reference. 
\begin{itemize}
    \item Clustering scenario (C3): both the APs and the MDCs are allocated based on the KMeans clustering. The resulting locations of these elements are depicted in Figure~\ref{img:edcs_aps_alloc_clus}. This Figure depicts the paths chosen by the pedestrians with points of different colors, corresponding with the coverage area of each MDCs. The different APs representing the mobile communication network are drawn as squares and the MDCs as circles. As a reference, public hospitals of this metropolitan area are marked with crosses. In this scenario we consider a fixed amount of 3 MDCs to illustrate the validity of the analysis.   
    \item Central hospital (H1): only one MDC is placed in the scenario, corresponding with the nearest hospital to its center. As a result,  all the tasks generated by the monitorization devices are sent to this single data center in this scenario.
    \item Nearest hospitals (H3): the MDCs of the clustering scenario (C3) are moved to the nearest hospitals. Figure~\ref{img:edcs_aps_alloc_hosp} represent the resulting scenario.
    \item All the hospitals (H9): an MDC is placed in each of the nine hospitals present in the selected metropolitan area.
\end{itemize}

As we aim to compare how the power consumption is distributed among the MDCs of these scenarios, we preserve the same configuration for all the MDCs. Each MDC is composed of 10 processing units (PUs), allocated in a single rack. Each of these PUs represents an Intel Core i9900K processor, operating at 3.6 GHz and composed of 8 cores (16 threads). For calculating  the power consumption of the PUs, we selected the \textit{IdleActive} power model included in the Mercury framework. This simple model considers two different power consumption values. The idle power consumption applies when no task is being executed in a specific PU. The active power consumption specifies the power consumption registered when the PU is executing one or more tasks. In this case, according to the selected processor, the idle power is set to 47W and the active power to 95W. 

Moreover, it is worth noting that the selected dispatching strategy searches the first PU sequentially with enough free resources to allocate new tasks. Therefore, a PU does not receive a task until the previous PU is unable to allocate it. Although Mercury allows configuring the shutdown of inactive processing units, all the PUs are powered on during all the simulations to reduce variability and facilitate the interpretation of the results.

For each pedestrian in the 10-hour dataset, an agent is created in the simulation. For this, Mercury sorts all the agents' entry time and creates and destroys them dynamically so that only the active agents are loaded in memory. We consider that all these pedestrian agents carry a monitoring device, and configure two types of services. The inference service requests periodically to the nearest MDC an updated estimation of the probability of a new onset of pain, based on the previously trained predictive models. The training service requests a new training of the patient model and aims to generate more efficient models with the updated patient data iteratively. 

\begin{table}[ht]
\caption{Statistics of the monitoring devices services.}
\label{tbl:mon_services}
\begin{tabular}{@{}llllll@{}}
\toprule
Service   & \begin{tabular}[c]{@{}l@{}}Generation\\ period\end{tabular} & \begin{tabular}[c]{@{}l@{}}PU Util.\\ (\%)\end{tabular} & \begin{tabular}[c]{@{}l@{}}Operation\\ time (s)\end{tabular} & \begin{tabular}[c]{@{}l@{}}Packet payload\\ size (B)\end{tabular} & \begin{tabular}[c]{@{}l@{}}Total packet\\ size (B)\end{tabular} \\
\midrule
Inference & 60s                                                         & 6.25                                                          & 1.17                                                         & 65                                                                & 119                                                             \\
Training  & unif(1s, 1d)                                                & 6.25                                                          & 18                                                           & 20                                                                & 74     \\
\bottomrule
\end{tabular}
\end{table}

Table~\ref{tbl:mon_services} shows more detailed information of these services. The inference service requests each minute an updated probability of pain onset, while the subsequent training request is calculated with a uniform distribution from 1 second to 1 day. Both services use 6.25\% of a PU, since they fully use one of its 16 threads.
The operation times for the inference and training services, 1.17s and 18s respectively, were extracted from actual training times of migraines models using similar processors. The packet payload size is determined according to the BSON encoding of the dictionaries exemplified in Figure~\ref{img:services_packets}. Both contain the patient identifier and a boolean indicating to the data center if new training is needed. Additionally, the inference packet contains the average values of the last minute for the implied hemodynamic variables. The total packet size of Table~\ref{tbl:mon_services} corresponds to the sum of the payload size and the header of the TCP packet (fixed to 54B).

\begin{table}[ht]
\caption{Mercury configuration parameters.}
\label{tbl:mercury_params}
\centering
\begin{tabular}{@{}ll@{}}
\toprule
Parameter & Value \\
\midrule
Mercury mode                                                                         & Lite                                                                  \\
Hot standby                                                                          & Off                                                                   \\
PUs per MDC                                                                          & 10                                                                    \\
Threads per PU                                                                       & 16                                                                    \\
Threads per service                                                                  & 1                                                                     \\
%Service start time                                                                   & 0.05s                                                                 \\
%Service stop time                                                                    & 0.05s                                                                 \\
\begin{tabular}[c]{@{}l@{}}Simultaneous requests\\ (per service and UE)\end{tabular} & 1                                                                     \\
Power model                                                                          & Idle/Active                                                           \\
Idle power                                                                           & 47W                                                                   \\
Active power                                                                         & 95W     \\
\bottomrule
\end{tabular}
\end{table}

\newsavebox{\inference}
\begin{lrbox}{\inference}% Store first listing
  \begin{lstlisting}[language=json,basicstyle=\small\ttfamily]
    {
      "id":348345,
      "train":false,
      "spo2":98,
      "eda":19.7,
      "temp":34.23,
      "hr":67
    }
\end{lstlisting}
\end{lrbox}

\newsavebox{\training}
\begin{lrbox}{\training}% Store first listing
  \begin{lstlisting}[language=json,basicstyle=\small\ttfamily]
    {
      "id":348345,
      "train":true
    }
  \end{lstlisting}
\end{lrbox}

\begin{figure}
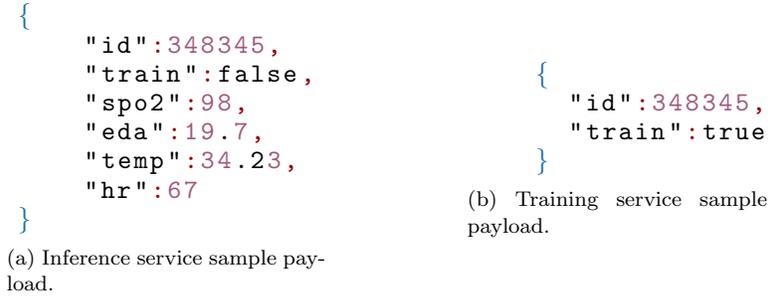

\centering
\subfloat[Inference service sample payload.] {
\begin{lstlisting}[language=json,basicstyle=\small\ttfamily]
{
    "id":348345,
    "train":false,
    "spo2":98,
    "eda":19.7,
    "temp":34.23,
    "hr":67
}
\end{lstlisting}  
  \label{img:services_packets_inference}
}
\hspace{50pt}
\subfloat[Training service sample payload.] {\usebox{\training}\label{img:services_packets_training}}
\caption{Sample packets generated by the simulation services.}
\label{img:services_packets}  
\end{figure}

Table~\ref{tbl:mercury_params} contains summary information regarding the specific configuration values used for this Mercury scenario. We used Mercury's Lite mode, which speeds up the execution time by avoiding fine-grained calculations regarding intermediate network communications. To improve the interpretability of the results, we do not active the hot standby configurations provided by Mercury, which allows dynamically change PUs status based on MDCs utilization ratios. Training and inference tasks are executed in a single thread, so 16 tasks can be run in parallel by each processing unit. Further information regarding the configuration of this scenario can be found in the project's official GitHub repository~\citep{github2021mdcs}, which includes the code and datasets used to deploy it.

Through this methodology, we can analyze the deployment of MDCs, studying how their location can alter the overall energy consumption of the scenario. In the next section, we present results regarding the optimization of energy consumption and the distribution of tasks for each of the mentioned MDCs organizations.

%%%%%%%%%%%%%%%%%%%%%%%%%%%%%%%%%%%%%%%%%%%%%%%%%%%%%%%%%%%%%%%%%%%%%%%%%%%%%%%%
%%%%%%%%%%%%%%%%%%%%%%%%%%%%%%%%%%%%%%%%%%%%%%%%%%%%%%%%%%%%%%%%%%%%%%%%%%%%%%%%

\section{Results}
\label{sec:results}

\begin{figure}[ht]
\centering
\includegraphics[width=1\textwidth]{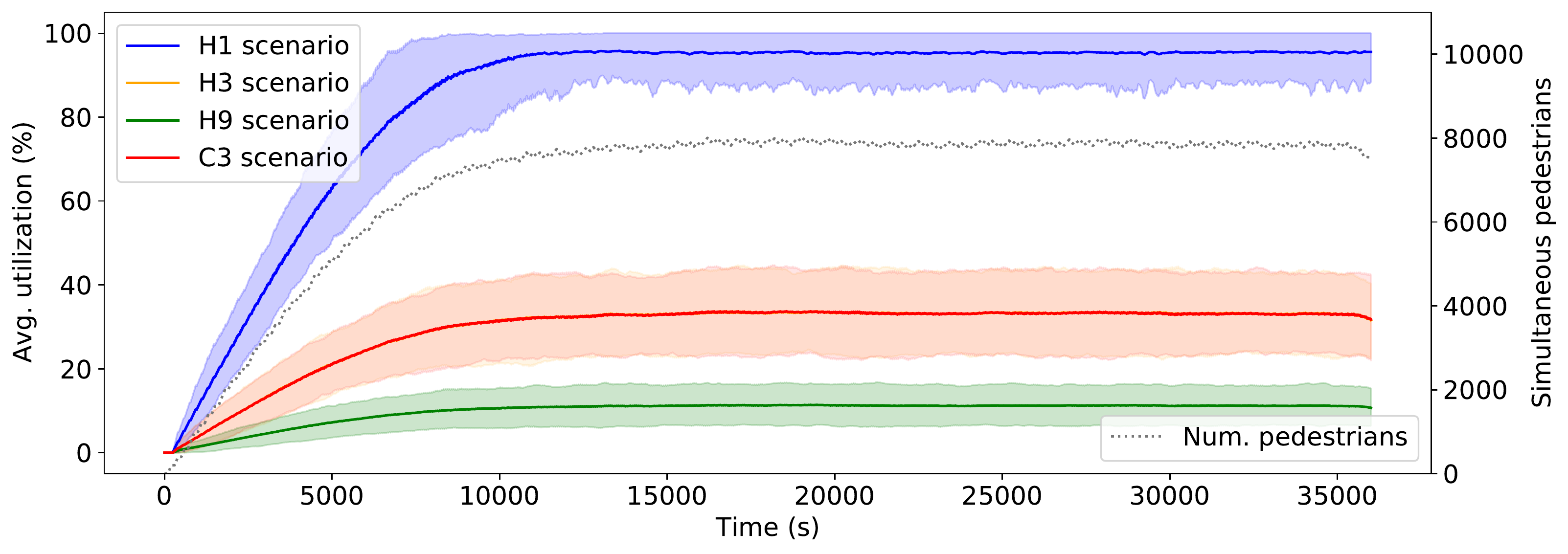}
\caption{MDCs average utilization for the different scenarios, compared with the number of simultaneous pedestrian agents. H3 and C3 produce similar results (overlapped).}
\label{img:scenarios_utilization_comparison}
\end{figure}

\begin{figure}[ht]
\centering
\includegraphics[width=1\textwidth]{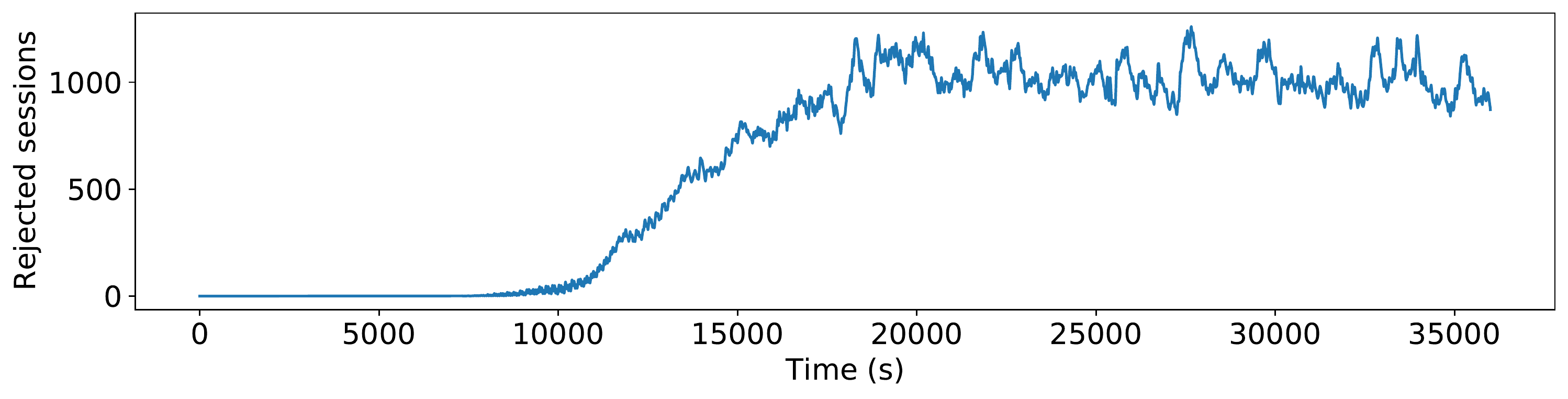}
\caption{Sessions rejected by the MDC of the H1 scenario due to resource overload .}
\label{img:h1_rejected_sessions}
\end{figure}

The four scenarios defined in the previous section were executed in the ETNA cluster of the Arcuitecture and Technology of Computing Systems (ArTeCS) research group of the Complutense University of Madrid. Each simulation run provides results like power consumption, utilization ratios, sessions status, etc. Figure~\ref{img:scenarios_utilization_comparison} shows how the average utilization of each MDCs evolves as simulation time progresses, and includes shaded areas whose edges indicate the minimum and maximum utilization over time. In this Figure, the solid lines represent the evolution of the utilization factor, and the dashed line represents the evolution in the number of simultaneously monitored pedestrians. We can see how the H1 hospital gets quickly overloaded, being unable to serve all incoming requests. This situation is depicted in Figure~\ref{img:h1_rejected_sessions}, showing its number of rejected sessions over time. H3 and C3 scenarios produce very similar results, being its evolution lines overlapped in the plot. Finally, because the requests are distributed among a greater number of MDCs, the H9 scenario obtains the least average utilization factor.

\begin{figure}
\centering
\subfloat[C3 scenario.] {
    \includegraphics[width=0.45\textwidth]{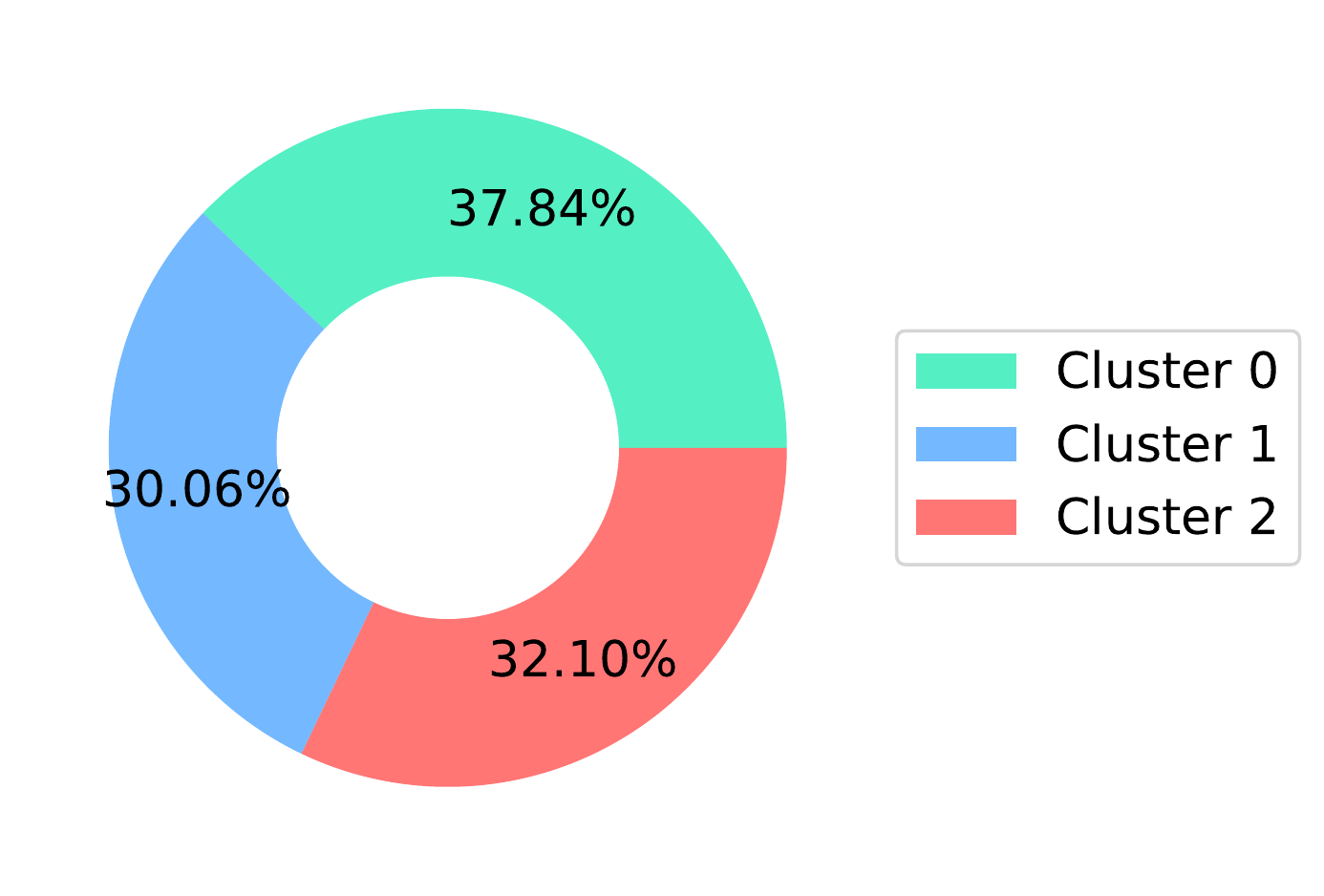}
    \label{img:pd_pie_clus3}
}
%\hspace{50pt}
\subfloat[H3 scenario.] {
    \includegraphics[width=0.45\textwidth]{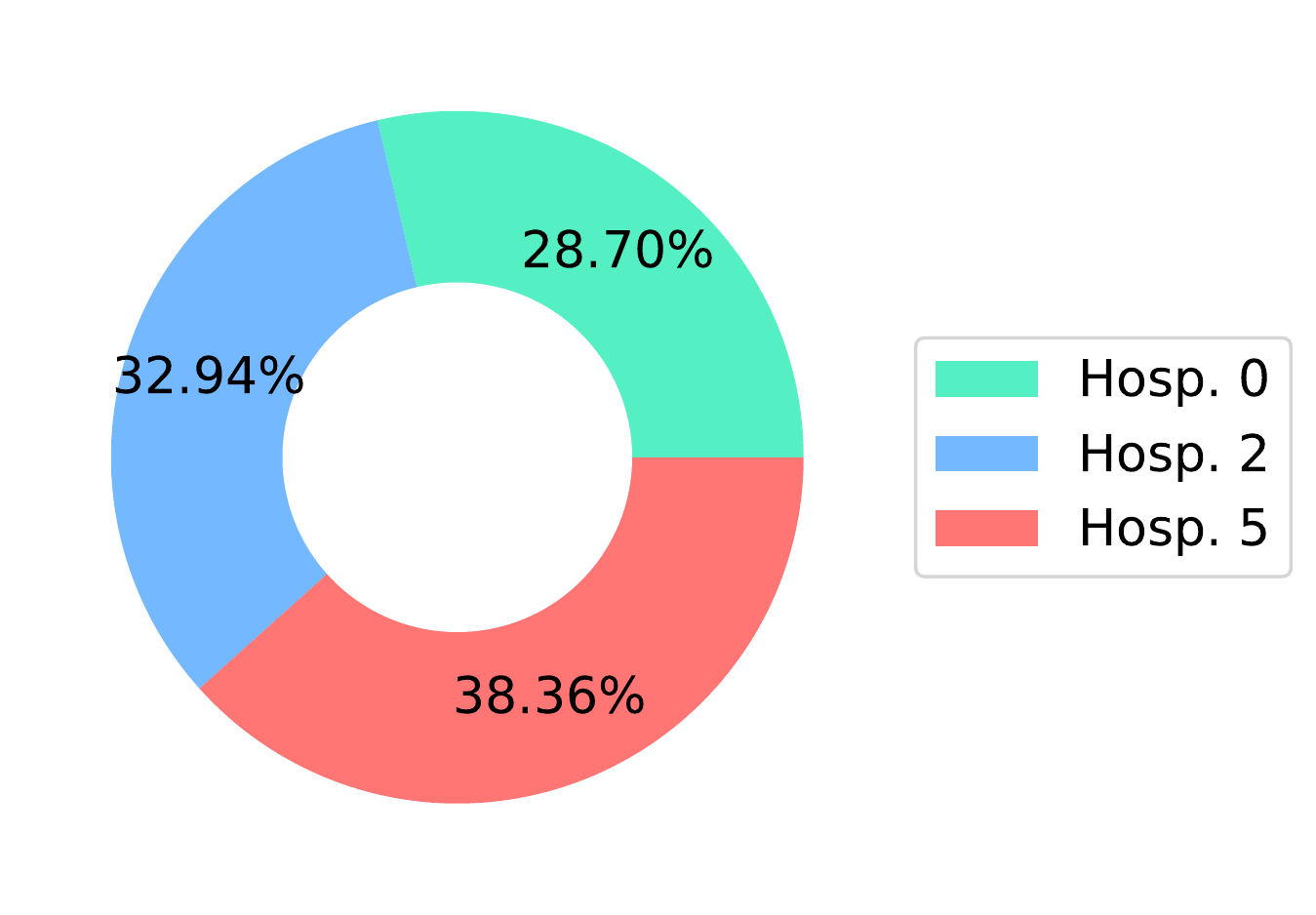}
    \label{img:pd_pie_hosp3}
}
\\
\subfloat[H9 scenario.] {
    \includegraphics[width=0.45\textwidth]{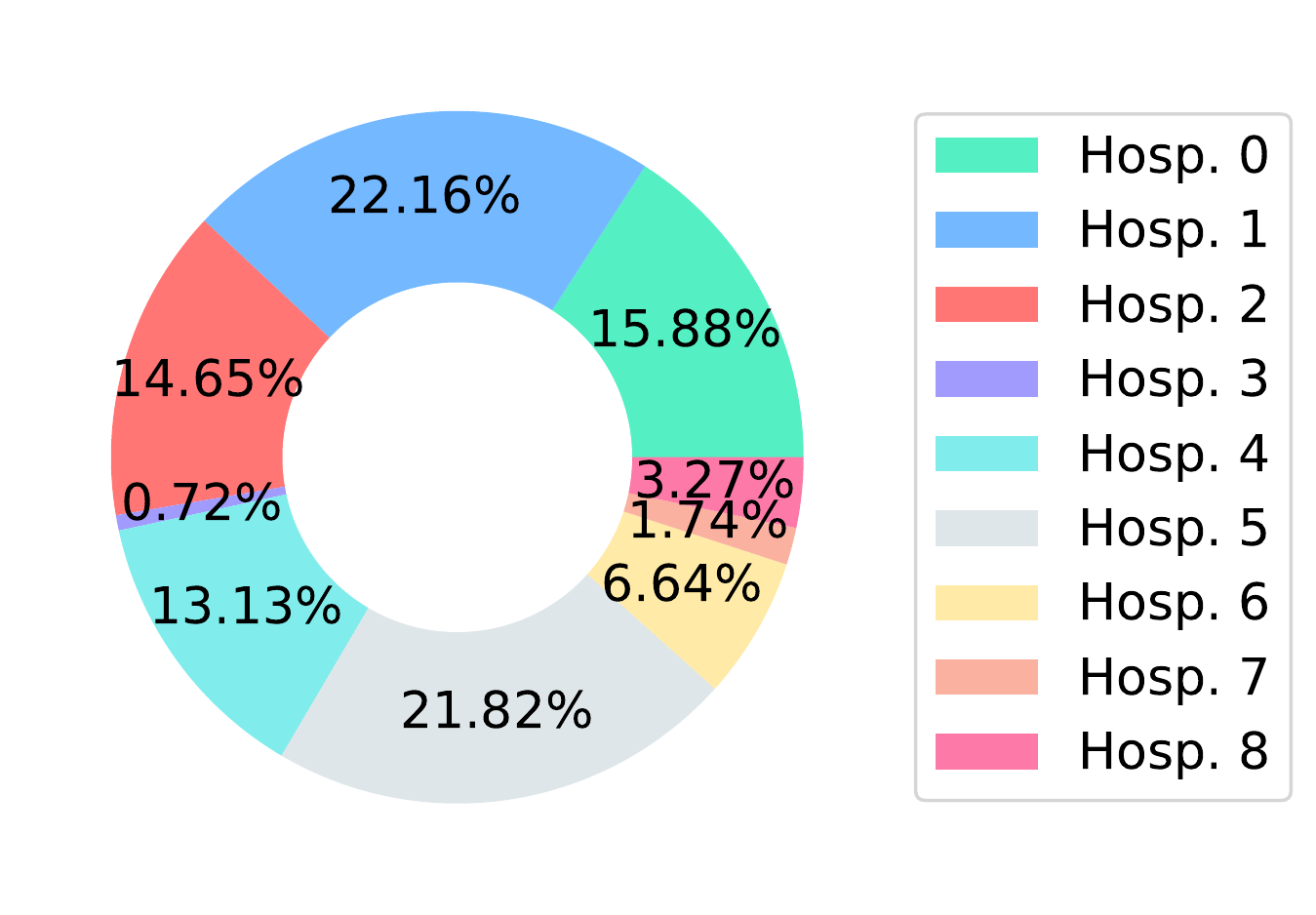}
    \label{img:pd_pie_hosp9}
}

\caption{Distribution of computational tasks over the Micro Data Centers.}
\label{img:pd_pies}  
\end{figure}

For a better understanding of how the requests are distributed over the MDCs, we represent  (see Figure~\ref{img:pd_pies}) the percentage of utilization of the individual MDCs with respect to the total utilization of each scenario. In the case of H3 and C3 scenarios, we can see an even distribution of the requests, sending about a third of them to each MDC. As expected, the percentage of individual usage differs considerably in the H9 scenario since the distance among data centers is not optimized. As a result, a third of data centers have less than 3\% usage while the two most used data centers add up to around 44\%.

% \begin{figure}
% \subfigure [Scenario 1: APs and MDCs are allocated using KMeans clustering.] {
%     \includegraphics[width=0.5\textwidth]{xl_scenario_clustered.pdf}
%     %\caption{Satellite view.}
%     \label{img:edcs_aps_alloc_clus}
% }
% \subfigure[Scenario 2: MDCs location are changed so that the correspond to the hospitals closest to the MDCs generated for Scenario 1. The same APs remain.]{
%     \includegraphics[width=0.5\textwidth]{xl_scenario_hosp_nearest.pdf}
%     \label{img:edcs_aps_alloc_hosp}
% }
% \caption{Scenarios considered in the simulation.}
% \label{img:edcs_aps_alloc}   
% \end{figure}

\begin{figure}[ht]
\centering
\includegraphics[width=0.6\textwidth]{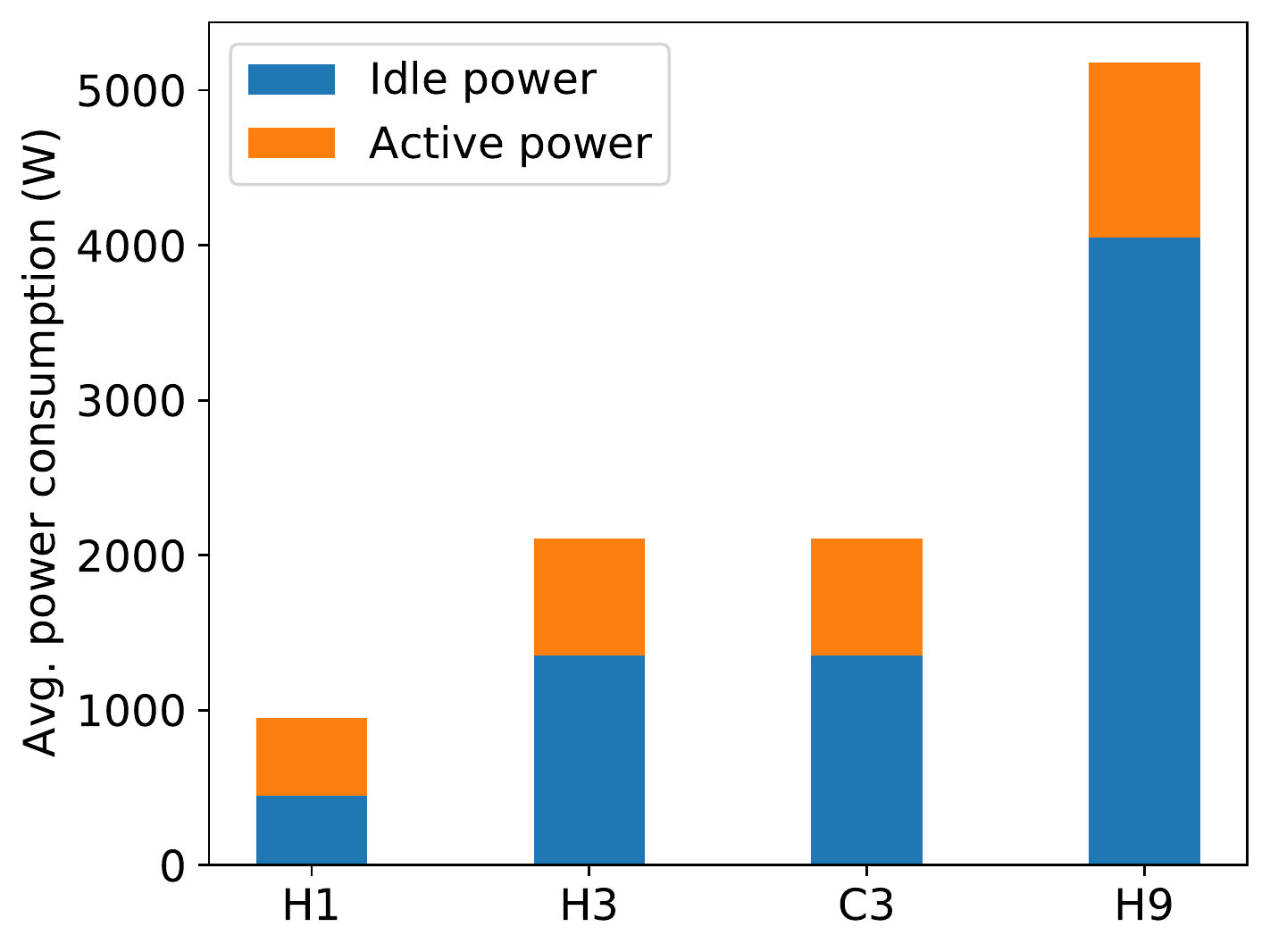}
\caption[Average power consumption]{Average power consumption of the different scenarios after the number of simultaneous pedestrians is stabilized (after 15000s of simulation time).}
\label{img:scenarios_power_comparison}
\end{figure}

Figure~\ref{img:scenarios_power_comparison} depicts the mean power consumption for the different MDCs after 15000s of simulation time, when the number of simultaneous pedestrians is stabilized. The blue bars indicate the power consumption derived from the infrastructure itself, while the orange ones  represent the consumption due to the training and inference services' execution. Consequently, we can see how H3 and C3 scenarios present an equivalent power consumption with the utilization ratios. Also, we can see how around a third of the power consumption corresponds to the execution of the services. While looking at the H1 scenario, it is worth noting that even though the energy consumption only represents around 47\% compared to the H3 and C3 scenarios, its MDC gets quickly overloaded before reaching a third of the total simulation time. Hence, this scenario does not provide a convenient service to the training and inference tasks and does not accomplish the scenario's requirements. The H9 scenario is not a good solution either, since it presents a total consumption almost 2.5 times higher than the scenarios with 3 MDCs. Moreover, as its processing units show a less exhaustive use, the resulting non-idle power consumption is 45\% more than the H3 and C3 scenarios.

%%%%%%%%%%%%%%%%%%%%%%%%%%%%%%%%%%%%%%%%%%%%%%%%%%%%%%%%%%%%%%%%%%%%%%%%%%%%%%%
%%%%%%%%%%%%%%%%%%%%%%%%%%%%%%%%%%%%%%%%%%%%%%%%%%%%%%%%%%%%%%%%%%%%%%%%%%%%%%%

\section{Conclusions}
\label{sec:conclusions}

We are in the midst of the data era, where the information generated worldwide and the number of connected devices keeps increasing exponentially. Such devices are helping to improve the efficiency of a large variety of procedures and activities. A highly relevant example is the field of healthcare, where non-intrusive monitoring devices are promoting a move from the traditional post-facto diagnose-and-treat paradigm to prognosis and predictive approaches. However, monitoring large populations with these networked devices demands vast storage and processing capabilities, together with efficient strategies for the deployment and management of the underlying IoT infrastructures. In this paper, we have presented a M\&S-driven methodology that can be used to analyze and deploy healthcare IoT infrastructures involving networks of monitored patients. We have simulated patients' movement using a well-known crowd simulator, over an urban scenario based on actual building layouts and metro stops. The architecture and functionality of data centers have been modeled with the use of the Mercury framework, defining specific APs over the city to serve as intermediate connection infrastructures for the communication between monitoring devices and data centers, and defining specific data center architectures and processing patterns. Also, we have compared different scenarios based on clustered and hospital-based Micro Data Centers locations, analyzing how the placement of the processing resources impacts the overall energy consumption of the system. As a result, we have obtained several comparisons of utilization ratios and power consumption, providing insights that can help in decision-making when optimizing the location of IoT infrastructures. Moreover, this methodology can be easily adapted to suit additional use cases, analyze other parameters as communication delays or suitability of the selected processing units, and perform an in-depth analysis of the optimal number of MDCs based on the chosen data center architectures.

\bibliographystyle{apacite}
\bibliography{biblio.bib}

\begin{thebibliography}{}

\bibitem [\protect \citeauthoryear {%
Aazam%
\ \BBA {} Huh%
}{%
Aazam%
\ \BBA {} Huh%
}{%
{\protect \APACyear {2015}}%
}]{%
aazam2015fog}
\APACinsertmetastar {%
aazam2015fog}%
\begin{APACrefauthors}%
Aazam, M.%
\BCBT {}\ \BBA {} Huh, E\BHBI N.%
\end{APACrefauthors}%
\unskip\
\newblock
\APACrefYearMonthDay{2015}{}{}.
\newblock
{\BBOQ}\APACrefatitle {Fog computing micro datacenter based dynamic resource
  estimation and pricing model for IoT} {Fog computing micro datacenter based
  dynamic resource estimation and pricing model for iot}.{\BBCQ}
\newblock
\BIn{} \APACrefbtitle {2015 IEEE 29th International Conference on Advanced
  Information Networking and Applications} {2015 ieee 29th international
  conference on advanced information networking and applications}\ (\BPGS\
  687--694).
\PrintBackRefs{\CurrentBib}

\bibitem [\protect \citeauthoryear {%
Afuang%
}{%
Afuang%
}{%
{\protect \APACyear {2020}}%
}]{%
idc2020iot}
\APACinsertmetastar {%
idc2020iot}%
\begin{APACrefauthors}%
Afuang, A.%
\end{APACrefauthors}%
\unskip\
\newblock
\APACrefYearMonthDay{2020}{}{}.
\newblock
\APACrefbtitle {{IoT} Growth Demands Rethink of Long-Term Storage Strategies}
  {{IoT} growth demands rethink of long-term storage strategies}\
  \APACbVolEdTR{}{\BTR{}}.
\newblock
\APACaddressInstitution{}{International Data Corporation}.
\PrintBackRefs{\CurrentBib}

\bibitem [\protect \citeauthoryear {%
Ali%
\ \protect \BOthers {.}}{%
Ali%
\ \protect \BOthers {.}}{%
{\protect \APACyear {2020}}%
}]{%
ali2020smart}
\APACinsertmetastar {%
ali2020smart}%
\begin{APACrefauthors}%
Ali, F.%
, El-Sappagh, S.%
, Islam, S\BPBI R.%
, Kwak, D.%
, Ali, A.%
, Imran, M.%
\BCBL {}\ \BBA {} Kwak, K\BHBI S.%
\end{APACrefauthors}%
\unskip\
\newblock
\APACrefYearMonthDay{2020}{}{}.
\newblock
{\BBOQ}\APACrefatitle {A smart healthcare monitoring system for heart disease
  prediction based on ensemble deep learning and feature fusion} {A smart
  healthcare monitoring system for heart disease prediction based on ensemble
  deep learning and feature fusion}.{\BBCQ}
\newblock
\APACjournalVolNumPages{Information Fusion}{63}{}{208--222}.
\PrintBackRefs{\CurrentBib}

\bibitem [\protect \citeauthoryear {%
\'Alvarez-S\'anchez%
}{%
\'Alvarez-S\'anchez%
}{%
{\protect \APACyear {2018}}%
}]{%
AlvarezSanchez2018}
\APACinsertmetastar {%
AlvarezSanchez2018}%
\begin{APACrefauthors}%
\'Alvarez-S\'anchez, C.%
\end{APACrefauthors}%
\unskip\
\newblock
\APACrefYearMonthDay{2018}{}{}.
\newblock
\APACrefbtitle {Deployment and management of platforms based on fog computing
  within an interconnected city} {Deployment and management of platforms based
  on fog computing within an interconnected city}\ \APACbVolEdTR{}{\BTR{}}.
\newblock
\APACaddressInstitution{}{Universidad Complutense de Madrid}.
\PrintBackRefs{\CurrentBib}

\bibitem [\protect \citeauthoryear {%
Barba%
, Mateos%
, Soto%
, Mezher%
\BCBL {}\ \BBA {} Igartua%
}{%
Barba%
\ \protect \BOthers {.}}{%
{\protect \APACyear {2012}}%
}]{%
barba2012smart}
\APACinsertmetastar {%
barba2012smart}%
\begin{APACrefauthors}%
Barba, C\BPBI T.%
, Mateos, M\BPBI A.%
, Soto, P\BPBI R.%
, Mezher, A\BPBI M.%
\BCBL {}\ \BBA {} Igartua, M\BPBI A.%
\end{APACrefauthors}%
\unskip\
\newblock
\APACrefYearMonthDay{2012}{}{}.
\newblock
{\BBOQ}\APACrefatitle {Smart city for VANETs using warning messages, traffic
  statistics and intelligent traffic lights} {Smart city for vanets using
  warning messages, traffic statistics and intelligent traffic lights}.{\BBCQ}
\newblock
\BIn{} \APACrefbtitle {2012 IEEE intelligent vehicles symposium} {2012 ieee
  intelligent vehicles symposium}\ (\BPGS\ 902--907).
\PrintBackRefs{\CurrentBib}

\bibitem [\protect \citeauthoryear {%
Behrisch%
, Bieker%
, Erdmann%
\BCBL {}\ \BBA {} Krajzewicz%
}{%
Behrisch%
\ \protect \BOthers {.}}{%
{\protect \APACyear {2011}}%
}]{%
behrisch2011sumo}
\APACinsertmetastar {%
behrisch2011sumo}%
\begin{APACrefauthors}%
Behrisch, M.%
, Bieker, L.%
, Erdmann, J.%
\BCBL {}\ \BBA {} Krajzewicz, D.%
\end{APACrefauthors}%
\unskip\
\newblock
\APACrefYearMonthDay{2011}{}{}.
\newblock
{\BBOQ}\APACrefatitle {SUMO--simulation of urban mobility: an overview}
  {Sumo--simulation of urban mobility: an overview}.{\BBCQ}
\newblock
\BIn{} \APACrefbtitle {Proceedings of SIMUL 2011, The Third International
  Conference on Advances in System Simulation.} {Proceedings of simul 2011, the
  third international conference on advances in system simulation.}
\PrintBackRefs{\CurrentBib}

\bibitem [\protect \citeauthoryear {%
Brogi%
\ \BBA {} Forti%
}{%
Brogi%
\ \BBA {} Forti%
}{%
{\protect \APACyear {2017}}%
}]{%
brogi2017fogtorch}
\APACinsertmetastar {%
brogi2017fogtorch}%
\begin{APACrefauthors}%
Brogi, A.%
\BCBT {}\ \BBA {} Forti, S.%
\end{APACrefauthors}%
\unskip\
\newblock
\APACrefYearMonthDay{2017}{}{}.
\newblock
{\BBOQ}\APACrefatitle {QoS-aware deployment of IoT applications through the
  fog} {Qos-aware deployment of iot applications through the fog}.{\BBCQ}
\newblock
\APACjournalVolNumPages{IEEE Internet of Things Journal}{4}{5}{1185--1192}.
\PrintBackRefs{\CurrentBib}

\bibitem [\protect \citeauthoryear {%
Calheiros%
, Ranjan%
, Beloglazov%
, De~Rose%
\BCBL {}\ \BBA {} Buyya%
}{%
Calheiros%
\ \protect \BOthers {.}}{%
{\protect \APACyear {2011}}%
}]{%
calheiros2011cloudsim}
\APACinsertmetastar {%
calheiros2011cloudsim}%
\begin{APACrefauthors}%
Calheiros, R\BPBI N.%
, Ranjan, R.%
, Beloglazov, A.%
, De~Rose, C\BPBI A.%
\BCBL {}\ \BBA {} Buyya, R.%
\end{APACrefauthors}%
\unskip\
\newblock
\APACrefYearMonthDay{2011}{}{}.
\newblock
{\BBOQ}\APACrefatitle {CloudSim: a toolkit for modeling and simulation of cloud
  computing environments and evaluation of resource provisioning algorithms}
  {Cloudsim: a toolkit for modeling and simulation of cloud computing
  environments and evaluation of resource provisioning algorithms}.{\BBCQ}
\newblock
\APACjournalVolNumPages{Software: Practice and experience}{41}{1}{23--50}.
\PrintBackRefs{\CurrentBib}

\bibitem [\protect \citeauthoryear {%
C{\'a}rdenas%
\ \protect \BOthers {.}}{%
C{\'a}rdenas%
\ \protect \BOthers {.}}{%
{\protect \APACyear {2020}}%
}]{%
cardenas2020mercury}
\APACinsertmetastar {%
cardenas2020mercury}%
\begin{APACrefauthors}%
C{\'a}rdenas, R.%
, Arroba, P.%
, Blanco, R.%
, Malag{\'o}n, P.%
, Risco-Mart{\'\i}n, J\BPBI L.%
\BCBL {}\ \BBA {} Moya, J\BPBI M.%
\end{APACrefauthors}%
\unskip\
\newblock
\APACrefYearMonthDay{2020}{}{}.
\newblock
{\BBOQ}\APACrefatitle {Mercury: A modeling, simulation, and optimization
  framework for data stream-oriented IoT applications} {Mercury: A modeling,
  simulation, and optimization framework for data stream-oriented iot
  applications}.{\BBCQ}
\newblock
\APACjournalVolNumPages{Simulation Modelling Practice and
  Theory}{101}{}{102037}.
\PrintBackRefs{\CurrentBib}

\bibitem [\protect \citeauthoryear {%
Castillo~O'Sullivan%
\ \BBA {} Thierer%
}{%
Castillo~O'Sullivan%
\ \BBA {} Thierer%
}{%
{\protect \APACyear {2015}}%
}]{%
castillo2015projecting}
\APACinsertmetastar {%
castillo2015projecting}%
\begin{APACrefauthors}%
Castillo~O'Sullivan, A.%
\BCBT {}\ \BBA {} Thierer, A\BPBI D.%
\end{APACrefauthors}%
\unskip\
\newblock
\APACrefYearMonthDay{2015}{}{}.
\newblock
{\BBOQ}\APACrefatitle {Projecting the growth and economic impact of the
  internet of things} {Projecting the growth and economic impact of the
  internet of things}.{\BBCQ}
\newblock
\APACjournalVolNumPages{Available at SSRN 2618794}{}{}{}.
\PrintBackRefs{\CurrentBib}

\bibitem [\protect \citeauthoryear {%
Dey%
, Abowd%
\BCBL {}\ \BBA {} Salber%
}{%
Dey%
\ \protect \BOthers {.}}{%
{\protect \APACyear {2000}}%
}]{%
dey2000context}
\APACinsertmetastar {%
dey2000context}%
\begin{APACrefauthors}%
Dey, A\BPBI K.%
, Abowd, G\BPBI D.%
\BCBL {}\ \BBA {} Salber, D.%
\end{APACrefauthors}%
\unskip\
\newblock
\APACrefYearMonthDay{2000}{}{}.
\newblock
{\BBOQ}\APACrefatitle {A context-based infrastructure for smart environments}
  {A context-based infrastructure for smart environments}.{\BBCQ}
\newblock
\BIn{} \APACrefbtitle {Managing Interactions in Smart Environments} {Managing
  interactions in smart environments}\ (\BPGS\ 114--128).
\newblock
\APACaddressPublisher{}{Springer}.
\PrintBackRefs{\CurrentBib}

\bibitem [\protect \citeauthoryear {%
Di%
\ \BBA {} Cappello%
}{%
Di%
\ \BBA {} Cappello%
}{%
{\protect \APACyear {2015}}%
}]{%
di2015gloudsim}
\APACinsertmetastar {%
di2015gloudsim}%
\begin{APACrefauthors}%
Di, S.%
\BCBT {}\ \BBA {} Cappello, F.%
\end{APACrefauthors}%
\unskip\
\newblock
\APACrefYearMonthDay{2015}{}{}.
\newblock
{\BBOQ}\APACrefatitle {GloudSim: Google trace based cloud simulator with
  virtual machines} {Gloudsim: Google trace based cloud simulator with virtual
  machines}.{\BBCQ}
\newblock
\APACjournalVolNumPages{Software: Practice and Experience}{45}{11}{1571--1590}.
\PrintBackRefs{\CurrentBib}

\bibitem [\protect \citeauthoryear {%
Erdemir%
\ \protect \BOthers {.}}{%
Erdemir%
\ \protect \BOthers {.}}{%
{\protect \APACyear {2020}}%
}]{%
erdemir2020credible}
\APACinsertmetastar {%
erdemir2020credible}%
\begin{APACrefauthors}%
Erdemir, A.%
, Mulugeta, L.%
, Ku, J\BPBI P.%
, Drach, A.%
, Horner, M.%
, Morrison, T\BPBI M.%
\BDBL {}Myers, J\BPBI G.%
\end{APACrefauthors}%
\unskip\
\newblock
\APACrefYearMonthDay{2020}{}{}.
\newblock
{\BBOQ}\APACrefatitle {Credible practice of modeling and simulation in
  healthcare: ten rules from a multidisciplinary perspective} {Credible
  practice of modeling and simulation in healthcare: ten rules from a
  multidisciplinary perspective}.{\BBCQ}
\newblock
\APACjournalVolNumPages{Journal of translational medicine}{18}{1}{1--18}.
\PrintBackRefs{\CurrentBib}

\bibitem [\protect \citeauthoryear {%
Fox%
\ \protect \BOthers {.}}{%
Fox%
\ \protect \BOthers {.}}{%
{\protect \APACyear {2009}}%
}]{%
fox2009above}
\APACinsertmetastar {%
fox2009above}%
\begin{APACrefauthors}%
Fox, A.%
, Griffith, R.%
, Joseph, A.%
, Katz, R.%
, Konwinski, A.%
, Lee, G.%
\BDBL {}others%
\end{APACrefauthors}%
\unskip\
\newblock
\APACrefYearMonthDay{2009}{}{}.
\newblock
{\BBOQ}\APACrefatitle {Above the clouds: A berkeley view of cloud computing}
  {Above the clouds: A berkeley view of cloud computing}.{\BBCQ}
\newblock
\APACjournalVolNumPages{Dept. Electrical Eng. and Comput. Sciences, University
  of California, Berkeley, Rep. UCB/EECS}{28}{13}{2009}.
\PrintBackRefs{\CurrentBib}

\bibitem [\protect \citeauthoryear {%
Gaba%
}{%
Gaba%
}{%
{\protect \APACyear {2007}}%
}]{%
gaba2007future}
\APACinsertmetastar {%
gaba2007future}%
\begin{APACrefauthors}%
Gaba, D\BPBI M.%
\end{APACrefauthors}%
\unskip\
\newblock
\APACrefYearMonthDay{2007}{}{}.
\newblock
{\BBOQ}\APACrefatitle {The future vision of simulation in healthcare} {The
  future vision of simulation in healthcare}.{\BBCQ}
\newblock
\APACjournalVolNumPages{Simulation in Healthcare}{2}{2}{126--135}.
\PrintBackRefs{\CurrentBib}

\bibitem [\protect \citeauthoryear {%
Giffin%
\ \protect \BOthers {.}}{%
Giffin%
\ \protect \BOthers {.}}{%
{\protect \APACyear {2003}}%
}]{%
giffin2003premonitory}
\APACinsertmetastar {%
giffin2003premonitory}%
\begin{APACrefauthors}%
Giffin, N.%
, Ruggiero, L.%
, Lipton, R\BPBI B.%
, Silberstein, S.%
, Tvedskov, J.%
, Olesen, J.%
\BDBL {}Macrae, A.%
\end{APACrefauthors}%
\unskip\
\newblock
\APACrefYearMonthDay{2003}{}{}.
\newblock
{\BBOQ}\APACrefatitle {Premonitory symptoms in migraine: an electronic diary
  study} {Premonitory symptoms in migraine: an electronic diary study}.{\BBCQ}
\newblock
\APACjournalVolNumPages{Neurology}{60}{6}{935--940}.
\PrintBackRefs{\CurrentBib}

\bibitem [\protect \citeauthoryear {%
Gupta%
, Vahid~Dastjerdi%
, Ghosh%
\BCBL {}\ \BBA {} Buyya%
}{%
Gupta%
\ \protect \BOthers {.}}{%
{\protect \APACyear {2017}}%
}]{%
gupta2017ifogsim}
\APACinsertmetastar {%
gupta2017ifogsim}%
\begin{APACrefauthors}%
Gupta, H.%
, Vahid~Dastjerdi, A.%
, Ghosh, S\BPBI K.%
\BCBL {}\ \BBA {} Buyya, R.%
\end{APACrefauthors}%
\unskip\
\newblock
\APACrefYearMonthDay{2017}{}{}.
\newblock
{\BBOQ}\APACrefatitle {iFogSim: A toolkit for modeling and simulation of
  resource management techniques in the Internet of Things, Edge and Fog
  computing environments} {ifogsim: A toolkit for modeling and simulation of
  resource management techniques in the internet of things, edge and fog
  computing environments}.{\BBCQ}
\newblock
\APACjournalVolNumPages{Software: Practice and Experience}{47}{9}{1275--1296}.
\PrintBackRefs{\CurrentBib}

\bibitem [\protect \citeauthoryear {%
Henares%
}{%
Henares%
}{%
{\protect \APACyear {{\protect \bibnodate {}}}}%
}]{%
github2021mdcs}
\APACinsertmetastar {%
github2021mdcs}%
\begin{APACrefauthors}%
Henares, K.%
\end{APACrefauthors}%
\unskip\
\newblock
\APACrefYearMonthDay{{\protect \bibnodate {}}}{}{}.
\newblock
\APACrefbtitle {{MDCs optimization scenario repository}.} {{MDCs optimization
  scenario repository}.}
\newblock
\APAChowpublished {\url{https://github.com/khvilaboa/mdcs_optimization}}.
\newblock
\APACrefnote{[Online; accessed 07-October-2021]}
\PrintBackRefs{\CurrentBib}

\bibitem [\protect \citeauthoryear {%
Jararweh%
\ \protect \BOthers {.}}{%
Jararweh%
\ \protect \BOthers {.}}{%
{\protect \APACyear {2017}}%
}]{%
jararweh2017software}
\APACinsertmetastar {%
jararweh2017software}%
\begin{APACrefauthors}%
Jararweh, Y.%
, Alsmirat, M.%
, Al-Ayyoub, M.%
, Benkhelifa, E.%
, Darabseh, A.%
, Gupta, B.%
\BCBL {}\ \BBA {} Doulat, A.%
\end{APACrefauthors}%
\unskip\
\newblock
\APACrefYearMonthDay{2017}{}{}.
\newblock
{\BBOQ}\APACrefatitle {Software-defined system support for enabling ubiquitous
  mobile edge computing} {Software-defined system support for enabling
  ubiquitous mobile edge computing}.{\BBCQ}
\newblock
\APACjournalVolNumPages{The Computer Journal}{60}{10}{1443--1457}.
\PrintBackRefs{\CurrentBib}

\bibitem [\protect \citeauthoryear {%
Kelman%
}{%
Kelman%
}{%
{\protect \APACyear {2006}}%
}]{%
kelman2006postdrome}
\APACinsertmetastar {%
kelman2006postdrome}%
\begin{APACrefauthors}%
Kelman, L.%
\end{APACrefauthors}%
\unskip\
\newblock
\APACrefYearMonthDay{2006}{}{}.
\newblock
{\BBOQ}\APACrefatitle {The postdrome of the acute migraine attack} {The
  postdrome of the acute migraine attack}.{\BBCQ}
\newblock
\APACjournalVolNumPages{Cephalalgia}{26}{2}{214--220}.
\PrintBackRefs{\CurrentBib}

\bibitem [\protect \citeauthoryear {%
Kolbe%
, Gr{\"o}ger%
\BCBL {}\ \BBA {} Pl{\"u}mer%
}{%
Kolbe%
\ \protect \BOthers {.}}{%
{\protect \APACyear {2005}}%
}]{%
kolbe2005citygml}
\APACinsertmetastar {%
kolbe2005citygml}%
\begin{APACrefauthors}%
Kolbe, T\BPBI H.%
, Gr{\"o}ger, G.%
\BCBL {}\ \BBA {} Pl{\"u}mer, L.%
\end{APACrefauthors}%
\unskip\
\newblock
\APACrefYearMonthDay{2005}{}{}.
\newblock
{\BBOQ}\APACrefatitle {CityGML: Interoperable access to 3D city models}
  {Citygml: Interoperable access to 3d city models}.{\BBCQ}
\newblock
\BIn{} \APACrefbtitle {Geo-information for disaster management}
  {Geo-information for disaster management}\ (\BPGS\ 883--899).
\newblock
\APACaddressPublisher{}{Springer}.
\PrintBackRefs{\CurrentBib}

\bibitem [\protect \citeauthoryear {%
Lera%
, Guerrero%
\BCBL {}\ \BBA {} Juiz%
}{%
Lera%
\ \protect \BOthers {.}}{%
{\protect \APACyear {2019}}%
}]{%
lera2019yafs}
\APACinsertmetastar {%
lera2019yafs}%
\begin{APACrefauthors}%
Lera, I.%
, Guerrero, C.%
\BCBL {}\ \BBA {} Juiz, C.%
\end{APACrefauthors}%
\unskip\
\newblock
\APACrefYearMonthDay{2019}{}{}.
\newblock
{\BBOQ}\APACrefatitle {YAFS: A simulator for IoT scenarios in fog computing}
  {Yafs: A simulator for iot scenarios in fog computing}.{\BBCQ}
\newblock
\APACjournalVolNumPages{IEEE Access}{7}{}{91745--91758}.
\PrintBackRefs{\CurrentBib}

\bibitem [\protect \citeauthoryear {%
Linde%
\ \protect \BOthers {.}}{%
Linde%
\ \protect \BOthers {.}}{%
{\protect \APACyear {2012}}%
}]{%
linde2012cost}
\APACinsertmetastar {%
linde2012cost}%
\begin{APACrefauthors}%
Linde, M.%
, Gustavsson, A.%
, Stovner, L\BPBI J.%
, Steiner, T\BPBI J.%
, Barr{\'e}, J.%
, Katsarava, Z.%
\BDBL {}others%
\end{APACrefauthors}%
\unskip\
\newblock
\APACrefYearMonthDay{2012}{}{}.
\newblock
{\BBOQ}\APACrefatitle {The cost of headache disorders in Europe: the Eurolight
  project} {The cost of headache disorders in europe: the eurolight
  project}.{\BBCQ}
\newblock
\APACjournalVolNumPages{European journal of neurology}{19}{5}{703--711}.
\PrintBackRefs{\CurrentBib}

\bibitem [\protect \citeauthoryear {%
Liu%
, Xu%
, Lu%
\BCBL {}\ \BBA {} Zhang%
}{%
Liu%
\ \protect \BOthers {.}}{%
{\protect \APACyear {2018}}%
}]{%
Liu2018}
\APACinsertmetastar {%
Liu2018}%
\begin{APACrefauthors}%
Liu, H.%
, Xu, B.%
, Lu, D.%
\BCBL {}\ \BBA {} Zhang, G.%
\end{APACrefauthors}%
\unskip\
\newblock
\APACrefYearMonthDay{2018}{}{}.
\newblock
{\BBOQ}\APACrefatitle {A path planning approach for crowd evacuation in
  buildings based on improved artificial bee colony algorithm} {A path planning
  approach for crowd evacuation in buildings based on improved artificial bee
  colony algorithm}.{\BBCQ}
\newblock
\APACjournalVolNumPages{Applied Soft Computing}{68}{}{360--376}.
\PrintBackRefs{\CurrentBib}

\bibitem [\protect \citeauthoryear {%
Masera%
, Bompard%
, Profumo%
\BCBL {}\ \BBA {} Hadjsaid%
}{%
Masera%
\ \protect \BOthers {.}}{%
{\protect \APACyear {2018}}%
}]{%
masera2018smart}
\APACinsertmetastar {%
masera2018smart}%
\begin{APACrefauthors}%
Masera, M.%
, Bompard, E\BPBI F.%
, Profumo, F.%
\BCBL {}\ \BBA {} Hadjsaid, N.%
\end{APACrefauthors}%
\unskip\
\newblock
\APACrefYearMonthDay{2018}{}{}.
\newblock
{\BBOQ}\APACrefatitle {Smart (electricity) grids for smart cities: Assessing
  roles and societal impacts} {Smart (electricity) grids for smart cities:
  Assessing roles and societal impacts}.{\BBCQ}
\newblock
\APACjournalVolNumPages{Proceedings of the IEEE}{106}{4}{613--625}.
\PrintBackRefs{\CurrentBib}

\bibitem [\protect \citeauthoryear {%
Mayer%
, Graser%
, Gupta%
, Saurez%
\BCBL {}\ \BBA {} Ramachandran%
}{%
Mayer%
\ \protect \BOthers {.}}{%
{\protect \APACyear {2017}}%
}]{%
mayer2017emufog}
\APACinsertmetastar {%
mayer2017emufog}%
\begin{APACrefauthors}%
Mayer, R.%
, Graser, L.%
, Gupta, H.%
, Saurez, E.%
\BCBL {}\ \BBA {} Ramachandran, U.%
\end{APACrefauthors}%
\unskip\
\newblock
\APACrefYearMonthDay{2017}{}{}.
\newblock
{\BBOQ}\APACrefatitle {Emufog: Extensible and scalable emulation of large-scale
  fog computing infrastructures} {Emufog: Extensible and scalable emulation of
  large-scale fog computing infrastructures}.{\BBCQ}
\newblock
\BIn{} \APACrefbtitle {2017 IEEE Fog World Congress (FWC)} {2017 ieee fog world
  congress (fwc)}\ (\BPGS\ 1--6).
\PrintBackRefs{\CurrentBib}

\bibitem [\protect \citeauthoryear {%
Medina%
, Lakhina%
, Matta%
\BCBL {}\ \BBA {} Byers%
}{%
Medina%
\ \protect \BOthers {.}}{%
{\protect \APACyear {2001}}%
}]{%
medina2001brite}
\APACinsertmetastar {%
medina2001brite}%
\begin{APACrefauthors}%
Medina, A.%
, Lakhina, A.%
, Matta, I.%
\BCBL {}\ \BBA {} Byers, J.%
\end{APACrefauthors}%
\unskip\
\newblock
\APACrefYearMonthDay{2001}{}{}.
\newblock
{\BBOQ}\APACrefatitle {BRITE: An approach to universal topology generation}
  {Brite: An approach to universal topology generation}.{\BBCQ}
\newblock
\BIn{} \APACrefbtitle {MASCOTS 2001, Proceedings Ninth International Symposium
  on Modeling, Analysis and Simulation of Computer and Telecommunication
  Systems} {Mascots 2001, proceedings ninth international symposium on
  modeling, analysis and simulation of computer and telecommunication systems}\
  (\BPGS\ 346--353).
\PrintBackRefs{\CurrentBib}

\bibitem [\protect \citeauthoryear {%
Mell%
\ \BBA {} Grance%
}{%
Mell%
\ \BBA {} Grance%
}{%
{\protect \APACyear {2011}}%
}]{%
mell2011nist}
\APACinsertmetastar {%
mell2011nist}%
\begin{APACrefauthors}%
Mell, P.%
\BCBT {}\ \BBA {} Grance, T.%
\end{APACrefauthors}%
\unskip\
\newblock
\APACrefYearMonthDay{2011}{}{}.
\newblock
\APACrefbtitle {The {NIST} definition of cloud computing} {The {NIST}
  definition of cloud computing}\ \APACbVolEdTR{}{\BTR{}}.
\newblock
\APACaddressInstitution{}{National Institute of Standards and Technology}.
\PrintBackRefs{\CurrentBib}

\bibitem [\protect \citeauthoryear {%
Mittal%
, Diallo%
\BCBL {}\ \BBA {} Tolk%
}{%
Mittal%
\ \protect \BOthers {.}}{%
{\protect \APACyear {2018}}%
}]{%
Mittal2018}
\APACinsertmetastar {%
Mittal2018}%
\begin{APACrefauthors}%
Mittal, S.%
, Diallo, S.%
\BCBL {}\ \BBA {} Tolk, A.%
\end{APACrefauthors}%
\unskip\
\newblock
\APACrefYear{2018}.
\newblock
\APACrefbtitle {Emergent behavior in complex systems engineering: a modeling
  and simulation approach} {Emergent behavior in complex systems engineering: a
  modeling and simulation approach}.
\newblock
\APACaddressPublisher{}{John Wiley \& Sons}.
\PrintBackRefs{\CurrentBib}

\bibitem [\protect \citeauthoryear {%
Mutlag%
, Abd~Ghani%
, Arunkumar%
, Mohammed%
\BCBL {}\ \BBA {} Mohd%
}{%
Mutlag%
\ \protect \BOthers {.}}{%
{\protect \APACyear {2019}}%
}]{%
mutlag2019enabling}
\APACinsertmetastar {%
mutlag2019enabling}%
\begin{APACrefauthors}%
Mutlag, A\BPBI A.%
, Abd~Ghani, M\BPBI K.%
, Arunkumar, N\BPBI a.%
, Mohammed, M\BPBI A.%
\BCBL {}\ \BBA {} Mohd, O.%
\end{APACrefauthors}%
\unskip\
\newblock
\APACrefYearMonthDay{2019}{}{}.
\newblock
{\BBOQ}\APACrefatitle {Enabling technologies for fog computing in healthcare
  IoT systems} {Enabling technologies for fog computing in healthcare iot
  systems}.{\BBCQ}
\newblock
\APACjournalVolNumPages{Future Generation Computer Systems}{90}{}{62--78}.
\PrintBackRefs{\CurrentBib}

\bibitem [\protect \citeauthoryear {%
N{\'u}{\~n}ez%
\ \protect \BOthers {.}}{%
N{\'u}{\~n}ez%
\ \protect \BOthers {.}}{%
{\protect \APACyear {2012}}%
}]{%
nunez2012icancloud}
\APACinsertmetastar {%
nunez2012icancloud}%
\begin{APACrefauthors}%
N{\'u}{\~n}ez, A.%
, V{\'a}zquez-Poletti, J\BPBI L.%
, Caminero, A\BPBI C.%
, Casta{\~n}{\'e}, G\BPBI G.%
, Carretero, J.%
\BCBL {}\ \BBA {} Llorente, I\BPBI M.%
\end{APACrefauthors}%
\unskip\
\newblock
\APACrefYearMonthDay{2012}{}{}.
\newblock
{\BBOQ}\APACrefatitle {iCanCloud: A flexible and scalable cloud infrastructure
  simulator} {icancloud: A flexible and scalable cloud infrastructure
  simulator}.{\BBCQ}
\newblock
\APACjournalVolNumPages{Journal of Grid Computing}{10}{1}{185--209}.
\PrintBackRefs{\CurrentBib}

\bibitem [\protect \citeauthoryear {%
Olesen%
}{%
Olesen%
}{%
{\protect \APACyear {2004}}%
}]{%
headache2004international}
\APACinsertmetastar {%
headache2004international}%
\begin{APACrefauthors}%
Olesen, J.%
\end{APACrefauthors}%
\unskip\
\newblock
\APACrefYearMonthDay{2004}{}{}.
\newblock
\APACrefbtitle {The international classification of headache disorders.} {The
  international classification of headache disorders.}
\newblock
\APAChowpublished {https://ichd-3.org/}.
\PrintBackRefs{\CurrentBib}

\bibitem [\protect \citeauthoryear {%
Pag{\'a}n%
\ \protect \BOthers {.}}{%
Pag{\'a}n%
\ \protect \BOthers {.}}{%
{\protect \APACyear {2015}}%
}]{%
pagan2015robust}
\APACinsertmetastar {%
pagan2015robust}%
\begin{APACrefauthors}%
Pag{\'a}n, J.%
, De~Orbe, M\BPBI I.%
, Gago, A.%
, Sobrado, M.%
, Risco-Mart{\'\i}n, J\BPBI L.%
, Mora, J\BPBI V.%
\BDBL {}Ayala, J\BPBI L.%
\end{APACrefauthors}%
\unskip\
\newblock
\APACrefYearMonthDay{2015}{}{}.
\newblock
{\BBOQ}\APACrefatitle {Robust and accurate modeling approaches for migraine
  per-patient prediction from ambulatory data} {Robust and accurate modeling
  approaches for migraine per-patient prediction from ambulatory data}.{\BBCQ}
\newblock
\APACjournalVolNumPages{Sensors}{15}{7}{15419--15442}.
\PrintBackRefs{\CurrentBib}

\bibitem [\protect \citeauthoryear {%
Pag{\'a}n%
, Risco-Mart{\'\i}n%
, Moya%
\BCBL {}\ \BBA {} Ayala%
}{%
Pag{\'a}n%
\ \protect \BOthers {.}}{%
{\protect \APACyear {2016}}%
}]{%
pagan2016grammatical}
\APACinsertmetastar {%
pagan2016grammatical}%
\begin{APACrefauthors}%
Pag{\'a}n, J.%
, Risco-Mart{\'\i}n, J\BPBI L.%
, Moya, J\BPBI M.%
\BCBL {}\ \BBA {} Ayala, J\BPBI L.%
\end{APACrefauthors}%
\unskip\
\newblock
\APACrefYearMonthDay{2016}{}{}.
\newblock
{\BBOQ}\APACrefatitle {Grammatical evolutionary techniques for prompt migraine
  prediction} {Grammatical evolutionary techniques for prompt migraine
  prediction}.{\BBCQ}
\newblock
\BIn{} \APACrefbtitle {Proceedings of the Genetic and Evolutionary Computation
  Conference 2016} {Proceedings of the genetic and evolutionary computation
  conference 2016}\ (\BPGS\ 973--980).
\PrintBackRefs{\CurrentBib}

\bibitem [\protect \citeauthoryear {%
Penas%
, Zapater%
, Risco-Mart{\'\i}n%
\BCBL {}\ \BBA {} Ayala%
}{%
Penas%
\ \protect \BOthers {.}}{%
{\protect \APACyear {2017}}%
}]{%
penas2017sfide}
\APACinsertmetastar {%
penas2017sfide}%
\begin{APACrefauthors}%
Penas, I.%
, Zapater, M.%
, Risco-Mart{\'\i}n, J\BPBI L.%
\BCBL {}\ \BBA {} Ayala, J\BPBI L.%
\end{APACrefauthors}%
\unskip\
\newblock
\APACrefYearMonthDay{2017}{}{}.
\newblock
{\BBOQ}\APACrefatitle {SFIDE: a simulation infrastructure for data centers}
  {Sfide: a simulation infrastructure for data centers}.{\BBCQ}
\newblock
\BIn{} \APACrefbtitle {Proceedings of the Summer Simulation Multi-Conference}
  {Proceedings of the summer simulation multi-conference}\ (\BPGS\ 1--12).
\PrintBackRefs{\CurrentBib}

\bibitem [\protect \citeauthoryear {%
Qayyum%
, Malik%
, Khattak%
, Khalid%
\BCBL {}\ \BBA {} Khan%
}{%
Qayyum%
\ \protect \BOthers {.}}{%
{\protect \APACyear {2018}}%
}]{%
qayyum2018fognetsim++}
\APACinsertmetastar {%
qayyum2018fognetsim++}%
\begin{APACrefauthors}%
Qayyum, T.%
, Malik, A\BPBI W.%
, Khattak, M\BPBI A\BPBI K.%
, Khalid, O.%
\BCBL {}\ \BBA {} Khan, S\BPBI U.%
\end{APACrefauthors}%
\unskip\
\newblock
\APACrefYearMonthDay{2018}{}{}.
\newblock
{\BBOQ}\APACrefatitle {FogNetSim++: A toolkit for modeling and simulation of
  distributed fog environment} {Fognetsim++: A toolkit for modeling and
  simulation of distributed fog environment}.{\BBCQ}
\newblock
\APACjournalVolNumPages{IEEE Access}{6}{}{63570--63583}.
\PrintBackRefs{\CurrentBib}

\bibitem [\protect \citeauthoryear {%
Reinsel%
, Gantz%
\BCBL {}\ \BBA {} Rydning%
}{%
Reinsel%
\ \protect \BOthers {.}}{%
{\protect \APACyear {2018}}%
}]{%
reinsel2018data}
\APACinsertmetastar {%
reinsel2018data}%
\begin{APACrefauthors}%
Reinsel, D.%
, Gantz, J.%
\BCBL {}\ \BBA {} Rydning, J.%
\end{APACrefauthors}%
\unskip\
\newblock
\APACrefYearMonthDay{2018}{}{}.
\newblock
{\BBOQ}\APACrefatitle {Data age 2025: the digitization of the world from edge
  to core} {Data age 2025: the digitization of the world from edge to
  core}.{\BBCQ}
\newblock
\APACjournalVolNumPages{Seagate,
  \url{https://www.seagate.com/files/www-content/our-story/trends/files/idc-seagate-dataage-whitepaper.pdf}}{}{}{}.
\PrintBackRefs{\CurrentBib}

\bibitem [\protect \citeauthoryear {%
Saha%
\ \protect \BOthers {.}}{%
Saha%
\ \protect \BOthers {.}}{%
{\protect \APACyear {2017}}%
}]{%
saha2017health}
\APACinsertmetastar {%
saha2017health}%
\begin{APACrefauthors}%
Saha, H\BPBI N.%
, Auddy, S.%
, Pal, S.%
, Kumar, S.%
, Pandey, S.%
, Singh, R.%
\BDBL {}Saha, S.%
\end{APACrefauthors}%
\unskip\
\newblock
\APACrefYearMonthDay{2017}{}{}.
\newblock
{\BBOQ}\APACrefatitle {Health monitoring using internet of things (IoT)}
  {Health monitoring using internet of things (iot)}.{\BBCQ}
\newblock
\BIn{} \APACrefbtitle {2017 8th Annual Industrial Automation and
  Electromechanical Engineering Conference (IEMECON)} {2017 8th annual
  industrial automation and electromechanical engineering conference
  (iemecon)}\ (\BPGS\ 69--73).
\PrintBackRefs{\CurrentBib}

\bibitem [\protect \citeauthoryear {%
Sahoo%
, Mohapatra%
\BCBL {}\ \BBA {} Wu%
}{%
Sahoo%
\ \protect \BOthers {.}}{%
{\protect \APACyear {2016}}%
}]{%
sahoo2016analyzing}
\APACinsertmetastar {%
sahoo2016analyzing}%
\begin{APACrefauthors}%
Sahoo, P\BPBI K.%
, Mohapatra, S\BPBI K.%
\BCBL {}\ \BBA {} Wu, S\BHBI L.%
\end{APACrefauthors}%
\unskip\
\newblock
\APACrefYearMonthDay{2016}{}{}.
\newblock
{\BBOQ}\APACrefatitle {Analyzing healthcare big data with prediction for future
  health condition} {Analyzing healthcare big data with prediction for future
  health condition}.{\BBCQ}
\newblock
\APACjournalVolNumPages{IEEE Access}{4}{}{9786--9799}.
\PrintBackRefs{\CurrentBib}

\bibitem [\protect \citeauthoryear {%
Schijve%
}{%
Schijve%
}{%
{\protect \APACyear {{\protect \bibnodate {}}}}%
}]{%
url2020pedestrians}
\APACinsertmetastar {%
url2020pedestrians}%
\begin{APACrefauthors}%
Schijve, L.%
\end{APACrefauthors}%
\unskip\
\newblock
\APACrefYearMonthDay{{\protect \bibnodate {}}}{}{}.
\newblock
\APACrefbtitle {{Pedestrian Dynamics}.} {{Pedestrian Dynamics}.}
\newblock
\APAChowpublished
  {\url{https://www.incontrolsim.com/software/pedestrian-dynamics/}}.
\newblock
\APACrefnote{[Online; accessed 25-November-2020]}
\PrintBackRefs{\CurrentBib}

\bibitem [\protect \citeauthoryear {%
Sonmez%
, Ozgovde%
\BCBL {}\ \BBA {} Ersoy%
}{%
Sonmez%
\ \protect \BOthers {.}}{%
{\protect \APACyear {2018}}%
}]{%
sonmez2018edgecloudsim}
\APACinsertmetastar {%
sonmez2018edgecloudsim}%
\begin{APACrefauthors}%
Sonmez, C.%
, Ozgovde, A.%
\BCBL {}\ \BBA {} Ersoy, C.%
\end{APACrefauthors}%
\unskip\
\newblock
\APACrefYearMonthDay{2018}{}{}.
\newblock
{\BBOQ}\APACrefatitle {Edgecloudsim: An environment for performance evaluation
  of edge computing systems} {Edgecloudsim: An environment for performance
  evaluation of edge computing systems}.{\BBCQ}
\newblock
\APACjournalVolNumPages{Transactions on Emerging Telecommunications
  Technologies}{29}{11}{e3493}.
\PrintBackRefs{\CurrentBib}

\bibitem [\protect \citeauthoryear {%
Stovner%
\ \BBA {} Andree%
}{%
Stovner%
\ \BBA {} Andree%
}{%
{\protect \APACyear {2010}}%
}]{%
stovner2010prevalence}
\APACinsertmetastar {%
stovner2010prevalence}%
\begin{APACrefauthors}%
Stovner, L\BPBI J.%
\BCBT {}\ \BBA {} Andree, C.%
\end{APACrefauthors}%
\unskip\
\newblock
\APACrefYearMonthDay{2010}{}{}.
\newblock
{\BBOQ}\APACrefatitle {Prevalence of headache in Europe: a review for the
  Eurolight project} {Prevalence of headache in europe: a review for the
  eurolight project}.{\BBCQ}
\newblock
\APACjournalVolNumPages{The journal of headache and pain}{11}{4}{289}.
\PrintBackRefs{\CurrentBib}

\bibitem [\protect \citeauthoryear {%
Tokody%
, Mezei%
\BCBL {}\ \BBA {} Schuster%
}{%
Tokody%
\ \protect \BOthers {.}}{%
{\protect \APACyear {2017}}%
}]{%
tokody2017overview}
\APACinsertmetastar {%
tokody2017overview}%
\begin{APACrefauthors}%
Tokody, D.%
, Mezei, I\BPBI J.%
\BCBL {}\ \BBA {} Schuster, G.%
\end{APACrefauthors}%
\unskip\
\newblock
\APACrefYearMonthDay{2017}{}{}.
\newblock
{\BBOQ}\APACrefatitle {An overview of autonomous intelligent vehicle systems}
  {An overview of autonomous intelligent vehicle systems}.{\BBCQ}
\newblock
\BIn{} \APACrefbtitle {Vehicle and Automotive Engineering} {Vehicle and
  automotive engineering}\ (\BPGS\ 287--307).
\newblock
\APACaddressPublisher{}{Springer}.
\PrintBackRefs{\CurrentBib}

\bibitem [\protect \citeauthoryear {%
Tripathi%
, Singh%
\BCBL {}\ \BBA {} Vishwakarma%
}{%
Tripathi%
\ \protect \BOthers {.}}{%
{\protect \APACyear {2019}}%
}]{%
Tripathi2019}
\APACinsertmetastar {%
Tripathi2019}%
\begin{APACrefauthors}%
Tripathi, G.%
, Singh, K.%
\BCBL {}\ \BBA {} Vishwakarma, D\BPBI K.%
\end{APACrefauthors}%
\unskip\
\newblock
\APACrefYearMonthDay{2019}{}{}.
\newblock
{\BBOQ}\APACrefatitle {Convolutional neural networks for crowd behaviour
  analysis: a survey} {Convolutional neural networks for crowd behaviour
  analysis: a survey}.{\BBCQ}
\newblock
\APACjournalVolNumPages{The Visual Computer}{35}{5}{753--776}.
\PrintBackRefs{\CurrentBib}

\bibitem [\protect \citeauthoryear {%
Tuli%
\ \protect \BOthers {.}}{%
Tuli%
\ \protect \BOthers {.}}{%
{\protect \APACyear {2020}}%
}]{%
tuli2020healthfog}
\APACinsertmetastar {%
tuli2020healthfog}%
\begin{APACrefauthors}%
Tuli, S.%
, Basumatary, N.%
, Gill, S\BPBI S.%
, Kahani, M.%
, Arya, R\BPBI C.%
, Wander, G\BPBI S.%
\BCBL {}\ \BBA {} Buyya, R.%
\end{APACrefauthors}%
\unskip\
\newblock
\APACrefYearMonthDay{2020}{}{}.
\newblock
{\BBOQ}\APACrefatitle {Healthfog: An ensemble deep learning based smart
  healthcare system for automatic diagnosis of heart diseases in integrated iot
  and fog computing environments} {Healthfog: An ensemble deep learning based
  smart healthcare system for automatic diagnosis of heart diseases in
  integrated iot and fog computing environments}.{\BBCQ}
\newblock
\APACjournalVolNumPages{Future Generation Computer Systems}{104}{}{187--200}.
\PrintBackRefs{\CurrentBib}

\bibitem [\protect \citeauthoryear {%
Voorsluys%
, Broberg%
\BCBL {}\ \BBA {} Buyya%
}{%
Voorsluys%
\ \protect \BOthers {.}}{%
{\protect \APACyear {2011}}%
}]{%
voorsluys2011introduction}
\APACinsertmetastar {%
voorsluys2011introduction}%
\begin{APACrefauthors}%
Voorsluys, W.%
, Broberg, J.%
\BCBL {}\ \BBA {} Buyya, R.%
\end{APACrefauthors}%
\unskip\
\newblock
\APACrefYearMonthDay{2011}{}{}.
\newblock
{\BBOQ}\APACrefatitle {Introduction to cloud computing} {Introduction to cloud
  computing}.{\BBCQ}
\newblock
\APACjournalVolNumPages{Cloud computing: Principles and paradigms}{}{}{1--44}.
\PrintBackRefs{\CurrentBib}

\bibitem [\protect \citeauthoryear {%
F.~Wang%
, Wang%
, Ma%
\BCBL {}\ \BBA {} Liu%
}{%
F.~Wang%
\ \protect \BOthers {.}}{%
{\protect \APACyear {2019}}%
}]{%
wang2019demystifying}
\APACinsertmetastar {%
wang2019demystifying}%
\begin{APACrefauthors}%
Wang, F.%
, Wang, F.%
, Ma, X.%
\BCBL {}\ \BBA {} Liu, J.%
\end{APACrefauthors}%
\unskip\
\newblock
\APACrefYearMonthDay{2019}{}{}.
\newblock
{\BBOQ}\APACrefatitle {Demystifying the crowd intelligence in last mile parcel
  delivery for smart cities} {Demystifying the crowd intelligence in last mile
  parcel delivery for smart cities}.{\BBCQ}
\newblock
\APACjournalVolNumPages{IEEE Network}{33}{2}{23--29}.
\PrintBackRefs{\CurrentBib}

\bibitem [\protect \citeauthoryear {%
M.~Wang%
, Li%
, Hu%
\BCBL {}\ \BBA {} Zhou%
}{%
M.~Wang%
\ \protect \BOthers {.}}{%
{\protect \APACyear {2013}}%
}]{%
url2020osm}
\APACinsertmetastar {%
url2020osm}%
\begin{APACrefauthors}%
Wang, M.%
, Li, Q.%
, Hu, Q.%
\BCBL {}\ \BBA {} Zhou, M.%
\end{APACrefauthors}%
\unskip\
\newblock
\APACrefYearMonthDay{2013}{}{}.
\newblock
{\BBOQ}\APACrefatitle {Quality analysis of open street map data} {Quality
  analysis of open street map data}.{\BBCQ}
\newblock
\APACjournalVolNumPages{International archives of the photogrammetry, remote
  sensing and spatial information sciences}{2}{}{}.
\PrintBackRefs{\CurrentBib}

\bibitem [\protect \citeauthoryear {%
Zeng%
\ \protect \BOthers {.}}{%
Zeng%
\ \protect \BOthers {.}}{%
{\protect \APACyear {2017}}%
}]{%
zeng2017iotsim}
\APACinsertmetastar {%
zeng2017iotsim}%
\begin{APACrefauthors}%
Zeng, X.%
, Garg, S\BPBI K.%
, Strazdins, P.%
, Jayaraman, P\BPBI P.%
, Georgakopoulos, D.%
\BCBL {}\ \BBA {} Ranjan, R.%
\end{APACrefauthors}%
\unskip\
\newblock
\APACrefYearMonthDay{2017}{}{}.
\newblock
{\BBOQ}\APACrefatitle {IOTSim: A simulator for analysing IoT applications}
  {Iotsim: A simulator for analysing iot applications}.{\BBCQ}
\newblock
\APACjournalVolNumPages{Journal of Systems Architecture}{72}{}{93--107}.
\PrintBackRefs{\CurrentBib}

\bibitem [\protect \citeauthoryear {%
B.~Zhou%
, Wang%
\BCBL {}\ \BBA {} Tang%
}{%
B.~Zhou%
\ \protect \BOthers {.}}{%
{\protect \APACyear {2012}}%
}]{%
Zhou2012}
\APACinsertmetastar {%
Zhou2012}%
\begin{APACrefauthors}%
Zhou, B.%
, Wang, X.%
\BCBL {}\ \BBA {} Tang, X.%
\end{APACrefauthors}%
\unskip\
\newblock
\APACrefYearMonthDay{2012}{}{}.
\newblock
{\BBOQ}\APACrefatitle {Understanding collective crowd behaviors: Learning a
  mixture model of dynamic pedestrian-agents} {Understanding collective crowd
  behaviors: Learning a mixture model of dynamic pedestrian-agents}.{\BBCQ}
\newblock
\BIn{} \APACrefbtitle {2012 IEEE Conference on Computer Vision and Pattern
  Recognition} {2012 ieee conference on computer vision and pattern
  recognition}\ (\BPGS\ 2871--2878).
\PrintBackRefs{\CurrentBib}

\bibitem [\protect \citeauthoryear {%
S.~Zhou%
\ \protect \BOthers {.}}{%
S.~Zhou%
\ \protect \BOthers {.}}{%
{\protect \APACyear {2010}}%
}]{%
zhou2010crowd}
\APACinsertmetastar {%
zhou2010crowd}%
\begin{APACrefauthors}%
Zhou, S.%
, Chen, D.%
, Cai, W.%
, Luo, L.%
, Low, M\BPBI Y\BPBI H.%
, Tian, F.%
\BDBL {}Hamilton, B\BPBI D.%
\end{APACrefauthors}%
\unskip\
\newblock
\APACrefYearMonthDay{2010}{}{}.
\newblock
{\BBOQ}\APACrefatitle {Crowd modeling and simulation technologies} {Crowd
  modeling and simulation technologies}.{\BBCQ}
\newblock
\APACjournalVolNumPages{ACM Transactions on Modeling and Computer Simulation
  (TOMACS)}{20}{4}{1--35}.
\PrintBackRefs{\CurrentBib}

\end{thebibliography}

\section*{Funding}
This work was partially supported by the Spanish Ministry of Science and Innovation under 2PIC4BioMed (PID2019-110866RB-I00) project grant, by the the Madrid Regional Government under CABAHLA-CM (S2018/TCS-4423) project grant, and by the Google Cloud Research Credits program with the award GCP19980904.

\end{document}